\documentclass[11pt, a4paper, logo, singlecolumn, copyright]{gensyn}
\usepackage{graphicx}

\usepackage{subcaption}
\usepackage{multirow}
\usepackage{amsmath}
\usepackage{rotating}
\usepackage[utf8]{inputenc} 
\usepackage[T1]{fontenc}    
\usepackage{hyperref}       
\usepackage{url}            
\usepackage{booktabs}       
\usepackage{amsfonts}       
\usepackage{nicefrac}       
\usepackage{microtype}      
\usepackage{xcolor}         
\usepackage{natbib}
\usepackage[english]{babel}

\newtheorem{theorem}{Theorem}
\newtheorem{lemma}{Lemma}
\newtheorem{conjecture}{Conjecture}

\newif\ifcomm
\commfalse

\usepackage{algorithm}
\usepackage{algpseudocode}
\usepackage{wrapfig}
\usepackage{booktabs}
\usepackage{multirow}

\title{NoLoCo: No-all-reduce Low Communication Training Method for Large Models}

\author[1]{Jari Kolehmainen}
\author[1]{Nikolay Blagoev}
\author[1]{Semih Kara}
\author[1]{John Donaghy}
\author[1]{Christopher Nies}
\author[1]{O\u{g}uzhan Ersoy}

\affil[1]{Gensyn}

\correspondingauthor{oguzhan@gensyn.ai}

\begin{abstract}
	Training large language models is generally done on clusters containing thousands of accelerators, communicating over a high-bandwidth interconnect. Scaling up these clusters is expensive and can become impractical, imposing limits on the size of models that can be trained. 
	Several recent studies have proposed training methods that are less communication intensive, avoiding the need for compute clusters with extremely high interconnect speeds. These low communication training methods still employ a global synchronization step for model parameters, which can be too costly with a high number of participants, as the communication cost scales quadratically with group size. In this work, we propose a novel optimization method, NoLoCo, that does not explicitly synchronize all model parameters during training and does not require any collective communication. NoLoCo implicitly synchronizes model weights via a novel variant of the Nesterov momentum optimizer by partially averaging model weights within randomly selected subgroups. We provide both a theoretical convergence analysis of our optimizer and empirical results from language model training.
	Our method requires significantly less communication than fully sharded data parallel training and DiLoCo, a widely used low-communication baseline. Moreover, our method avoids global blocking communication, thereby reducing accelerator idle time. Our experiments show that NoLoCo is more communication-efficient than DiLoCo, improving final perplexity by up to $4\%$ and converging up to $4\times$ faster in wall-clock time across a range of worker counts, model sizes, and communication bandwidths.
\end{abstract}

\begin{document}

\maketitle

\section{Introduction}

Large language models (LLMs) have recently shown impressive performance on a wide variety of tasks, including natural language understanding~\citep{liu2024deepseek,team2023gemini,touvron2023llama,zhang2022opt}; image related tasks~\citep{zhang2025llava,gao2024clip,zhu2023minigpt,lin2023video}; or speech recognition and generation~\citep{maiti2024voxtlm, gourav2024multi,rubenstein2023audiopalm,xu2025qwen2}. These large models are usually trained by a combination of different distributed training methods such as data parallelism~\citep{rasley2020deepspeed}, pipeline parallelism~\citep{huang2019gpipe,narayanan2021efficient,sun2024seq1f1b}, and others~\citep{shoeybi2019megatron,rasley2020deepspeed,liu2023ring,shyam2024tree,fujii2024accelerating,liu2024deepseek,cai2024survey}. These training methods are bandwidth-intensive and require high-bandwidth interconnects between compute nodes, which are typically available only in dedicated machine learning clusters~\citep{team2023gemini,grattafiori2024llama,duan2024efficient,intelligence2024amazon}. This requirement increases the cost of training and poses a limit on the training scale as highly connected computer clusters cannot be scaled easily beyond a data center.

Recently, a number of studies have aimed to address this issue by proposing algorithms that scale better than traditional distributed training algorithms in low-bandwidth and high latency networks~\citep{douillard2023diloco,douillard2024dipaco,ryabinin2023swarm,li2022branch,biswas2024boosting,kale2025eager,charles2025communication}. Most of these methods use an explicit step to synchronize the data parallel instances of the model during training, typically using all-reduce operations. This synchronization step can take several minutes in highly distributed networks and lead to poor overall scaling efficiency of the training algorithm~\citep{jaghouar2024intellect,yuan2022decentralized}.

\paragraph{Contributions.} 
The main contribution of this paper is NoLoCo: a novel optimization method that eliminates the need for all-to-all communication. NoLoCo combines the inner-outer optimizer paradigm with the idea of epidemic learning, replacing global synchronization with averaging among smaller subsets of accelerators. Each inner step additionally uses random pipeline routing, which implicitly aligns the different pipelines without explicit synchronization. To keep the weights at each pipeline stage from drifting apart, we modify the Nesterov momentum optimizer and prove convergence of the modified scheme.

On the task of pre-training a language model, we compare NoLoCo against traditional distributed training, the well-established low-communication baseline DiLoCo~\citep{douillard2023diloco}, and a range of recent gossip-like and asynchronous training methods. We use two datasets (Pushshift Reddit and C4) and multiple model sizes (125M, 1.3B, and 6.8B parameters). Our experiments show that NoLoCo is more communication-efficient than DiLoCo, improving final perplexity by up to $4\%$ and converging up to $4\times$ faster in wall-clock time across a range of worker counts, model sizes, and communication bandwidths. The speed-up from omitting the all-to-all communication increases with increasing number of workers and network latency variance.
Source code for running the experiments is available in \href{https://github.com/gensyn-ai/noloco}{GitHub}. 

\section{Related Work}
\label{sec:relatedwork}

\paragraph{Decentralized training methods.}
Standard data-parallel training keeps all workers synchronized by all-reducing gradients before every optimizer step~\citep{rasley2020deepspeed}. Decentralized training methods such as DiLoCo relax this assumption by allowing the model parameters to diverge during local steps and only synchronizing them at the outer optimizer steps, performed less frequently~\citep{douillard2024dipaco,li2022branch,biswas2024boosting,kale2025eager,demo,charles2025communication,qi2025dilocox, sarfi2025communication}. These methods split optimization into inner and outer steps, similar to lookahead optimizers~\citep{zhang2019lookahead}: workers update local weights independently during the inner steps, and synchronize via an outer gradient that is applied to update global weights shared between all copies of the model~\citep{douillard2023diloco}. This dramatically reduces all-reduce frequency, but the outer step can still dominate wall-clock time when interconnects are slow or worker counts are large~\citep{yuan2022decentralized}. Recent work mitigates
this by overlapping communication with compute~\citep{kale2025eager,qi2025dilocox,douillard2025streaming} or sparsifying it~\citep{sarfi2025communication}; both of these ideas can be extended to NoLoCo to further reduce the communication overhead.

A complementary line of work replaces global all-reduce with local or sub-group communication~\citep{hegedHus2019gossip,de2023epidemic,du2024hyperprism,ghiasvand2025decentralized}. Epidemic learning~\citep{de2023epidemic}, for example, synchronizes only a subset of weights via local averaging, building on classical gossip-based consensus methods~\citep{blot2016gossip,daily2018gossip}. These approaches are closely related to decentralized averaging and gossip-based methods~\citep{blot2016gossip,daily2018gossip}, which achieve global consensus by having the workers average their parameters with a peer.

\paragraph{Dynamic pipeline routing.}
Pipeline parallelism~\citep{huang2019gpipe,narayanan2021efficient,sun2024seq1f1b} is a popular choice for model parallelism in low bandwidth environments, as it requires less network bandwidth and involves less blocking communication than fully sharded data parallel (FSDP) training with ZeRO-3~\citep{rasley2020deepspeed} or tensor parallelism~\citep{shoeybi2019megatron}. In pipeline parallel training, the model is split in several stages, each stage holding a consecutive subset of layers. During the forward pass, model stages must wait for inputs from the preceding stage, and during the backward pass, for gradients from the subsequent stage. This leads to formation of a computation bubble where certain devices will be inherently idle

SWARM~\citep{ryabinin2023swarm} reduces blocking communication and improves load balancing by allowing each pipeline stage to receive inputs from any replica of the preceding stage; with equal workers and uniform topology its routing reduces to random routing. Subsequent work has explored more
efficient ad-hoc pipeline formations~\citep{blagoev2025flow}, while DiPaCo~\citep{douillard2024dipaco} replaces random routing with a learned sample-level router loosely inspired by mixture-of-experts parallelism~\citep{cai2024survey}. We adopt random routing in this work for simplicity; the other routing schemes above are orthogonal to NoLoCo and can be combined with it.

\section{NoLoCo}
\label{sec:noloco}

NoLoCo utilizes Data Parallelism (DP) and Pipeline Parallelism (PP) with the following modifications: (i) during the inner optimization steps, activations are routed through a randomly chosen pipeline path at each iteration rather than a fixed one, (ii) the outer optimization step synchronizes only pairs of accelerators instead of performing a global all-reduce across the data-parallel group. To prevent weights at the same pipeline stage from drifting apart across accelerators, we modify the Nesterov momentum optimizer step.

\subsection{Inner optimizer step via dynamic pipeline routing}

\begin{figure*}
    \centering
    \begin{subfigure}[t]{0.4\textwidth}
    \includegraphics[width=\linewidth]{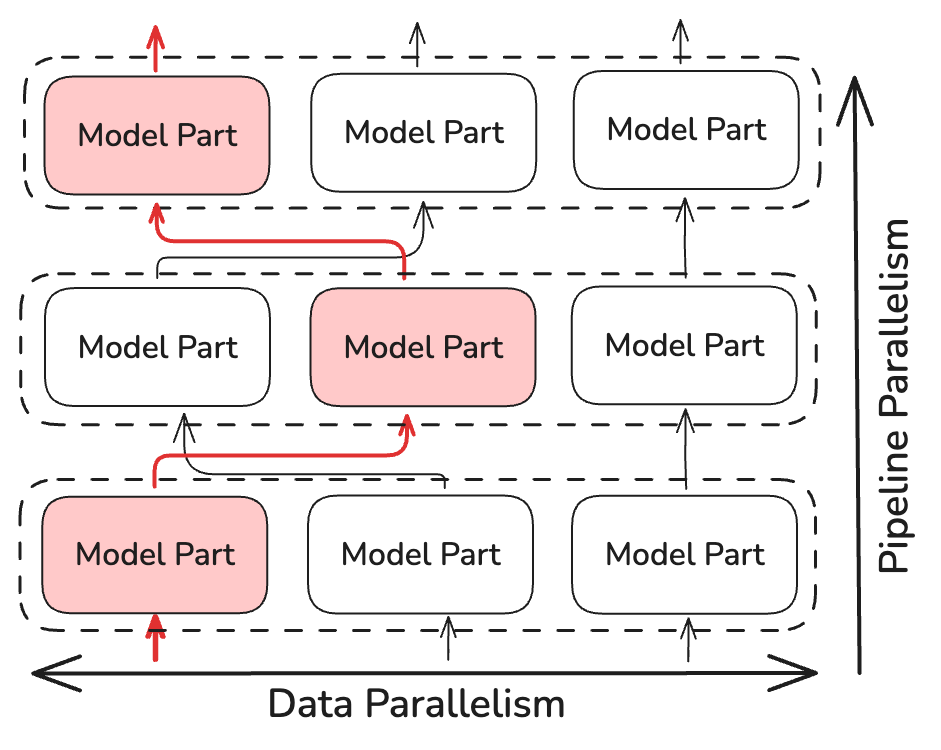}
    \caption{}
    \label{fig:illustration:A}
    \end{subfigure}
    \hspace{0.2in}
    \begin{subfigure}[t]{0.4\textwidth}
    \includegraphics[width=\linewidth]{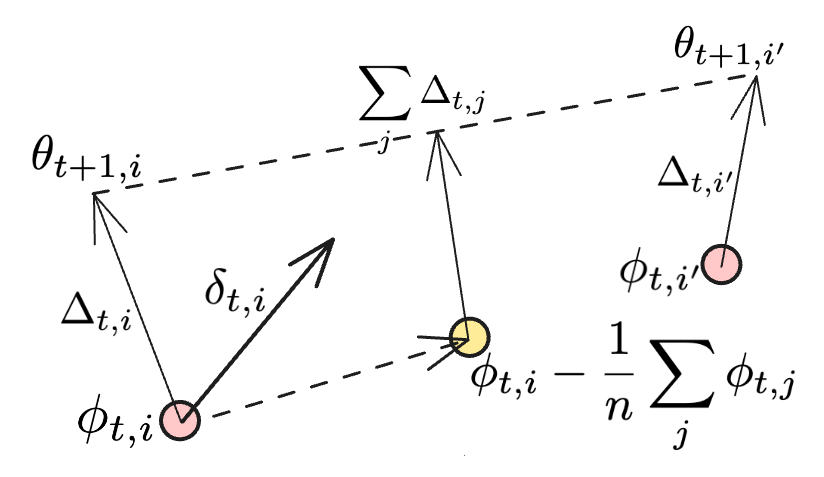}
    \caption{}
    \label{fig:illustration:B}
    \end{subfigure}
    \caption{$(a)$: Illustration of dynamic PP routing with DP. Model is split to consecutive pipeline stages (shown vertically), and each stage is replicated to process data in parallel (shown horizontally). $(b)$: Illustration of different terms of the outer momentum term for a group consisting of two workers. Red dots show the current slow weights. Yellow dot shows the average of the slow weights.
    }
    \label{fig:illustration}
\end{figure*}

NoLoCo uses dynamic PP, where an accelerator can receive inputs from any instance of the prior pipeline stage and forwards outputs to any instance of the next pipeline stage. 
At each step, we assign workers to groups via random permutations and route within these groups. Re-randomizing every step balances communication load across accelerators. This is illustrated in Fig. \ref{fig:illustration:A}. Gradients flow back along the same pipeline path used in the forward pass.

Because the routing changes each step, every pipeline stage interacts with activations and gradients from many different DP replicas over the course of training. This implicitly mixes information across replicas, reducing weight divergence without an explicit all-reduce. We hypothesize this creates an implicit effect to drive the weights of different DP models closer without requiring frequent synchronization. We give a pseudocode for the inner optimizer step with dynamic pipeline routing in Algorithm~\ref{alg:random_routing} (under Appendix~\ref{app:pp_routing}).

\subsection{Outer optimizer step with modified Nesterov momentum}
We relax the outer synchronisation by synchronizing the model parameters within a smaller local group, rather than the all-to-all reduction performed in other local-SGD methods.

Let $N$ denote the number of model instances in the whole network. We synchronize among a randomly chosen local subgroup of $n$ instances where $N \gg n$. We re-sample the subgroups every iteration so workers exchange information with different peers over time. In the local subgroup, we have $n$ model instances at step $t$, $\phi_{t,i}$, where $i$ denotes the model instance. We refer to $\phi_{t,i}$ as slow weights as in the Lookahead optimizer~\citep{zhang2019lookahead}. We then run $m$ local optimizer steps on $\phi_{t,i}$ to produce the fast weights $\theta_{t+1,i}$. This step is identical to the inner optimizer step of DiLoCo and resembles the inner steps of the lookahead optimizer. We use the fast model weights to compute a local outer gradient:
\begin{equation}
\Delta_{t,i} = \theta_{t+1,i} - \phi_{t,i}.
\label{eq:outer_gradient}
\end{equation} 
We modify the expression for Nesterov momentum by computing it over a local group as opposed to all data parallel workers, and introduce a term to account for the difference in the local slow weights:
\begin{equation}
\label{eq:delta}
\delta_{t,i} = \alpha \delta_{t-1,i} - \frac{\beta}{n} \left( \sum_j {\Delta_{t,j}} \right) - \gamma \left(\phi_{t,i} - \frac{1}{n} \sum_j \phi_{t,j}\right),
\end{equation} 
where $\alpha$ is the Nesterov momentum parameter; $\beta$ is the outer learning rate; $\gamma$ is a local weight averaging parameter. Finally, we update the local weights by the momentum $\delta_{t,i}$ in the same way as in Lookahead optimizer:
\begin{equation}
\phi_{t+1, i} = \phi_{t, i} + \delta_{t, i}.
\label{eq:update}
\end{equation} 
Fig. \ref{fig:illustration:B} illustrates the relationships between different terms in the method for outer gradient computation involving two workers.

If the subgroup consists of all model instances, then Eq.~\ref{eq:delta} simplifies to regular DiLoCo outer optimizer momentum and the last term vanishes: $\delta_{t} = \alpha \delta_{t-1} - \beta/n ( \textstyle\sum_j {\Delta_{t,j}})$. The third term in Eq.~\ref{eq:delta} can be viewed as a rolling average of weight differences between the worker and randomly sampled peers. 

In our experiments, we use the minimum subgroup size, which is $n=2$ (we present an ablation study on $n$ in Section~\ref{sec:n_ablation}). During the outer optimizer step, workers have to share their outer gradients (Eq. \ref{eq:outer_gradient}) and prior slow weights $\phi_{t,i}$. The prior slow weights can be communicated already at the end of the prior outer step. This allows overlapping communication of the slow weights with the computation for the next fast weights. 

Intuitively, the method should have convergence properties very similar to those of DiLoCo as the mean term in Eq. \ref{eq:delta} is a rolling average of the slow weights over the duration of training. We make this precise by showing that the modified Nesterov optimizer in Eq.~\ref{eq:delta} converges to the optimum $\theta=0$ for a quadratic loss \( \mathcal{L}(\theta) = \frac{1}{2}(\theta-c)^{\mathrm{T}} \mathrm{A} (\theta-c)\), where \(c\sim\mathcal{N}(0,\Sigma)\) with a constant covariance matrix \(\Sigma\), and \(\mathrm{A}\) is a symmetric positive definite matrix\footnote{These assumptions (quadratic objective, positive definite curvature, and normally distributed shifts in the optimum) are standard in analyses of momentum and accelerated methods, and are widely adopted in prior work~\citep{schaul2013no,wu2018understanding,zhang2019lookahead}.}. We also assume that the inner optimizer uses stochastic gradient descent with a constant learning rate $\omega$. With these assumptions following theorem holds:
\begin{theorem}
    As the outer iteration count $t$ tends to infinity, the slow weights satisfy $\mathbb{E}(\phi_{t,i}) \to 0$ and $\mathbb{V}(\phi_{t,i}) \propto \omega^2$ for all $i=1,\dots,N$.
    \label{theorem:convergence}
\end{theorem}

Thus the models converge to a neighbourhood of the optimal weights; we present the proof in Appendix \ref{section::convergence_analsis}. The variance bound also suggests that the inner learning rate (and its schedule in particular) provides an effective handle on weight divergence during training: starting with a large learning rate and decaying it towards the end yields a tight cluster of models. We present empirical evidence for this in Section~\ref{sec:results}.

\section{Results and Discussion}
\label{sec:results}

\subsection{Experiment setup}
\label{sec:exp_setting}

We focus on language modeling with a next-token prediction objective. We use two datasets, pushshift Reddit~\citep{baumgartner2020pushshift} and C4~\citep{dodge2021documenting}, to probe the training approaches. For benchmarking, we use $10$ million tokens held out from the training data for Reddit and the validation partition for C4. All data is tokenized by the Llama sentencepiece tokenizer with a vocabulary size of $128,000$ tokens and formatted to sequences of $1024$ tokens. We explore 3 model sizes: small, medium  and large Llama models \citep{grattafiori2024llama3} with 125M, 1.3B and 6.8B parameters respectively, details of which can be found in Table \ref{table:model_parameters}.

We take DiLoCo as our primary baseline and compare against it in detail. We also report comparisons with gossip-like and asynchronous training methods (specifically Epidemic Learning \citep{de2023epidemic}, SparseLoCo \citep{sarfi2025communication}, PalSGD \citep{naganuma2025pseudoasync}, Async LocSGD \citep{liu2024asynclocalsgd} and GWTF \citep{blagoev2025flow}) in Appendix~\ref{appendix:lit_comparison}. 

In all experiments, we use Adam as the (inner) optimizer, clip gradients to \(1.0\) and set the outer learning rate $\beta=0.7$. For DiLoCo we use a momentum value of $\alpha=0.3$ (which we found to produce better results in our setting than the standard $\alpha=0.9$) and perform outer optimization at every $H=100$ inner steps. For NoLoCo, we use a higher momentum value of $\alpha=0.5$; a group size of two workers; and perform the outer optimizer step every at $50$ inner steps\footnote{It is worth noting that even with the doubled frequency of outer optimizer steps, \textit{NoLoCo still requires much less communication than DiLoCo} since $N > 2\cdot n =4$.}. Additional experiment details are in Appendix~\ref{Appx:hyper}. We report hyperparameter ablations for NoLoCo in Section~\ref{sec:hparam_ablations}.

\subsection{Perplexity results}

Table \ref{tab:result} displays validation perplexities (PPL) at the end of the training for Reddit and C4 datasets. We observe that both NoLoCo and DiLoCo are slightly worse than fully sharded data parallel (FSDP) training, typically few percent worse, but in some cases even $30\%$ (C4, Small model, 16 data parallel world size). The PPL gap is generally larger with smaller models and large data parallel world sizes; and decreases when the model size becomes larger or the data parallel world size becomes smaller. This finding is consistent with the findings of a recent study on DiLoCo scaling~\citep{charles2025communication}. In this regard, NoLoCo's convergence scales with data-parallel world size and model size in much the same way as DiLoCo's.

NoLoCo achieves better final perplexity than DiLoCo in most experiments (across all Reddit runs and the majority of C4 runs). This seems counter-intuitive, as one would expect the variations in model weights during training to slow convergence, not accelerate it. We hypothesize that this is due to small perturbations in model weights acting as a regularizer. All training in this study stays within the first epoch, but large datasets are known to contain near-duplicate sequences that can induce mild over-fitting even within a single epoch. This may also explain why the effect is more pronounced on Reddit than on C4, since C4's broader topical diversity makes such over-fitting less likely.

\begin{table*}[]
    \centering
    \caption{Final validation PPL for different world sizes and models. DP stands for the data parallel world size and PP for the number of pipeline stages. FSDP stands for fully sharded data parallel training. Lowest PPL among the low-communication methods are highlighted with a bold font.}
    \label{tab:result}
    \begin{tabular}{l | r r r || r | r r || r | r r}
    & & & & \multicolumn{3}{c||}{Pushshift Reddit} & \multicolumn{3}{|c}{C4} \\
    Model & Total & DP & PP & FSDP & DiLoCo & NoLoCo & FSDP & DiLoCo & NoLoCo \\
    \hline
    \multirow{4}{*}{Small}
    & $8$ & $8$ & $1$ & $25.5$ & $27.6$ & $\bold{27.3}$ & $29.1$ & $35.4$ & $\bold{34.5}$\\
    & $8$ & $4$ & $2$ & $25.7$ & $26.8$ & $\bold{26.4}$ & $29.1$ & $32.1$ & $\bold{31.3}$ \\
    & $16$ & $8$ & $2$ & $25.5$ & $27.6$ & $\bold{27.2}$ & $29.1$ & $34.0$ & $\bold{33.1}$ \\
    & $32$ & $16$ & $2$ & $25.5$ & $29.7$ & $\bold{29.1}$ & $29.1$ & $39.1$ & $\bold{37.7}$ \\
    \hline
    \hline
    \multirow{3}{*}{Medium}
    & $16$ & $8$ & $2$ & $19.6$ & $21.0$ & $\bold{20.5}$ & $18.8$ & $21.8$ & $\bold{21.1}$ \\
    & $32$ & $16$ & $2$ & $19.6$ & $21.2$ & $\bold{20.7}$ & $18.8$ & $\bold{23.2}$ & $23.4$ \\
    & $64$ & $16$ & $4$ & $19.6$ & $21.0$ & $\bold{20.9}$ & $18.8$ & $\bold{22.6}$ & $22.9$ \\
    \hline
    \hline
    \multirow{1}{*}{Large} & $64$ & $16$ & $4$ & $16.1$ & $18.0$ & $\bold{17.5}$ & $15.7$ & $17.3$ & $\bold{16.6}$
    \end{tabular}
\end{table*}

\begin{figure}[ht]
    \centering
    \hspace{-0.15in}
    \begin{subfigure}[t]{0.33\linewidth}
        \centering
        \includegraphics[height=4cm]{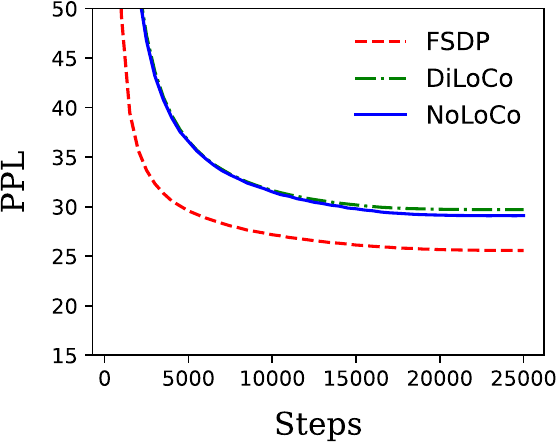}
        \caption{\hspace*{-1.3cm}}
        \label{fig:ppl_results:A}
    \end{subfigure}
    \hfill
    \begin{subfigure}[t]{0.33\linewidth}
        \centering
        \includegraphics[height=4cm]{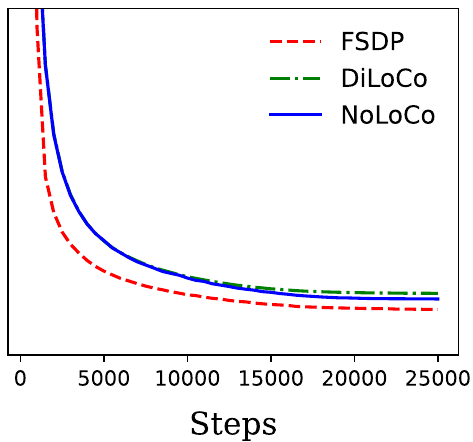}
        \caption{}
        \label{fig:ppl_results:B}
    \end{subfigure}
    \hspace{-0.1in}
    \begin{subfigure}[t]{0.33\linewidth}
        \centering
        \includegraphics[height=4cm]{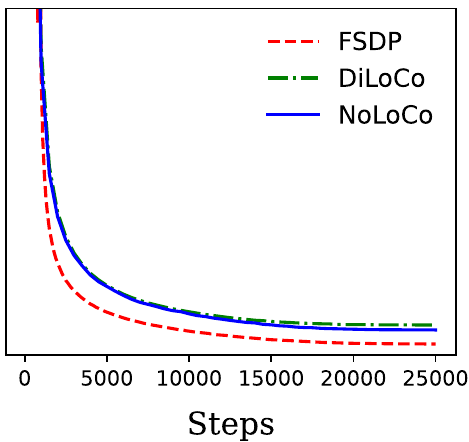}
        \caption{}
        \label{fig:ppl_results:C}
    \end{subfigure}
    \caption{Reddit validation PPL of different training methods at different optimizer step counts. Solid blue curve is NoLoCo, dashed red curve show FSDP, and dashed green curve is DiLoCo. $(a)$: Small model with DP world size of $8$ and two pipeline stages; $(b)$: Medium model with DP world size of $8$ and two pipeline stages. $(c)$: Large model with DP world size of $16$ and four pipeline stages.
    }
    \label{fig:ppl_results}
\end{figure}

\begin{figure}[t]
    \centering
    \begin{subfigure}[t]{0.495\linewidth}
        \centering
        \includegraphics[height=4.5cm]{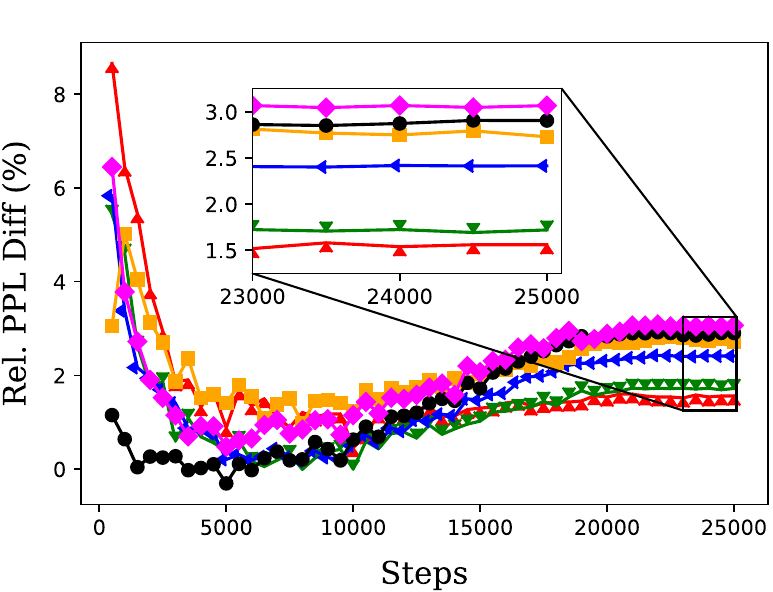}
        \caption{\hspace*{-0.7cm}}
        \label{fig:train_results:A}
    \end{subfigure}
    \begin{subfigure}[t]{0.495\linewidth}
        \centering
        \includegraphics[height=4.5cm]{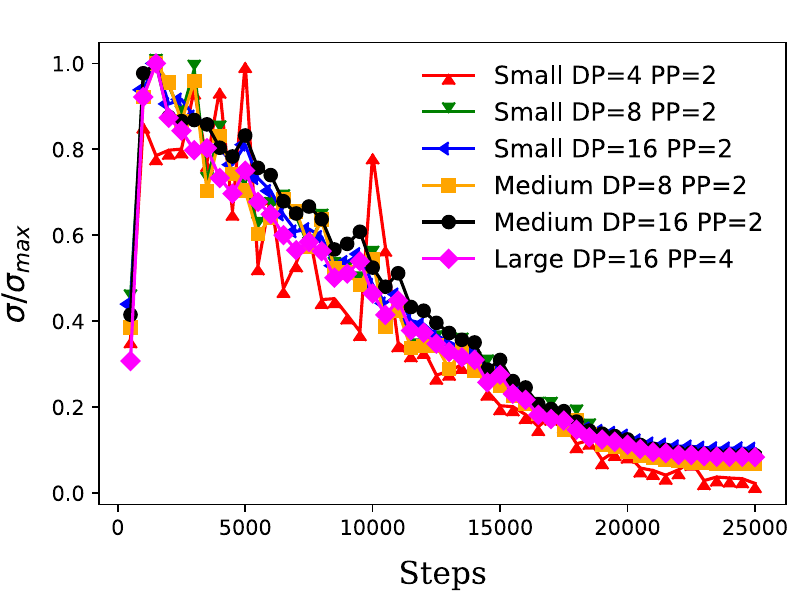}
        \caption{\hspace*{-1cm}}
        \label{fig:train_results:B}
    \end{subfigure}
    \caption{$(a)$: Relative validation PPL difference between DiLoCo and NoLoCo. Perplexity numbers are normalized by the FSDP PPL at the same optimizer step count. Positive number indicate faster convergence compared to DiLoCo. $(b)$: Standard deviation of the model weights across the data parallel world size, normalized by the largest value within the training run.
    }
    \label{fig:result2}
\end{figure}

Fig. \ref{fig:ppl_results} displays validation PPL over the course of training for all model sizes (see Appendix \ref{appendix:training_loss} for the corresponding training loss). We observe that the gap between FSDP and the decentralized methods decrease as the model size grows, and NoLoCo attains lower PPL than DiLoCo towards the end of training. This is most clearly seen in Fig.~\ref{fig:train_results:A}, which plots the relative validation PPL difference between DiLoCo and NoLoCo on Reddit, defined as
\begin{equation*}
\text{Relative PPL Diff} = \frac{\text{DiLoCo PPL} - \text{NoLoCo PPL}}{\text{FSDP PPL}}.
\end{equation*}
We see that NoLoCo typically achieves a few percent lower PPL early and late in training, with the two methods tracking closely in between. The late-stage gap is dataset-dependent: on Reddit, NoLoCo converges faster towards the end, whereas on C4 the outcome varies with model size and data-parallel world size.

Finally, Fig.~\ref{fig:train_results:B} shows how much the replicas' weights disagree over the course of training. The spread grows during early training, peaking near the end of warm-up when the learning rate is highest and replicas drift apart fastest, then shrinks as the learning rate decays and the replicas reconverge. This matches Theorem~\ref{theorem:convergence}, which predicts that each replica's weights have variance proportional to the square of the inner learning rate near convergence, and hence that the inter-replica standard deviation should track the learning rate linearly. Empirically, the Pearson correlation coefficient of the standard deviation and the learning rate indeed ranges from $0.91$ to $0.97$, confirming that the inner learning rate is the dominant control on weight divergence. Therefore, learning rate scheduling can be used effectively to obtain eventual consistency of the weights in NoLoCo.

\subsection{Latency results}
\label{section::latency}

\paragraph{Throughput analysis.}
To measure NoLoCo's wall-clock performance in realistic settings, we recorded the average inner-step times for FSDP, DiLoCo, and NoLoCo on 32 workers under \textit{bandwidth-limited interconnects}. Each run uses the medium model with pure DP, and executes two outer steps using a different interconnect speeds. 

The slowest interlink speed ($100\mathrm{Mb/s}$) is characteristic for regular consumer grade internet; the middle speed ($1\mathrm{Gb/s}$) can be found in high grade consumer internet connections; and $100\mathrm{Gb/s}$ represents interlink speeds available in data centers.
Table~\ref{tab:latency} reports the resulting average inner training times. It can be seen that, even at $100\,\mathrm{Gb/s}$, FSDP is much slower than both decentralized methods. Table~\ref{tab:latency} also reports throughput in inner steps per second, computed as $\mathrm{Throughput} = H / T_{\text{outer}}$, where $H$ is the number of inner steps per outer step.

\begin{table}[t]
\centering
\caption{Wall-clock time per inner step and corresponding throughput in inner steps per second for all methods at varying GPU interlink bandwidths. All runs use 32 GPUs, medium model size, 100 inner steps per outer step, two outer steps total, and no PP.}
\label{tab:latency}
\begin{tabular}{l rr rr rr}
\multirow{2}{*}{Method}
& \multicolumn{2}{c}{$100\,\mathrm{Mb/s}$}
& \multicolumn{2}{c}{$1\,\mathrm{Gb/s}$}
& \multicolumn{2}{c}{$100\,\mathrm{Gb/s}$} \\
\cmidrule(lr){2-3}
\cmidrule(lr){4-5}
\cmidrule(lr){6-7}
& Time (s) & Throughput
& Time (s) & Throughput
& Time (s) & Throughput \\
\midrule
FSDP
& 5368.41 & 0.019
& 548.97  & 0.182
& 41.39   & 2.416 \\
DiLoCo
& 61.80 & 1.618
& 6.55  & 15.267
& 1.70  & 58.824 \\
NoLoCo
& \textbf{15.26} & \textbf{6.553}
& \textbf{2.80}  & \textbf{35.714}
& \textbf{1.64}  & \textbf{60.976}
\end{tabular}
\end{table}

\paragraph{Wall-clock time analysis.} 
While NoLoCo's per-iteration convergence is only marginally faster than DiLoCo's (up to $4\%$), its wall-clock advantage under bandwidth constraints is substantial. As shown in Table~\ref{tab:latency}, at a consumer-grade Wi-Fi bandwidth of $100\,\mathrm{Mb/s}$ NoLoCo is $4\times$ faster than DiLoCo and $351\times$ faster than FSDP per inner step. The corresponding speed-ups at $1\,\mathrm{Gb/s}$ are still substantial ($2.34\times$ for DiLoCo and $196\times$ for FSDP).

Combined with the comparable per-iteration convergence reported in Section~\ref{sec:results}, this implies that NoLoCo can reach a target validation PPL nearly $4\times$ faster than DiLoCo in wall-clock time on consumer-grade networks.

\begin{figure}[t]
    \centering
    \begin{subfigure}[t]{0.32\textwidth}
        \centering
        \includegraphics[width=\linewidth]{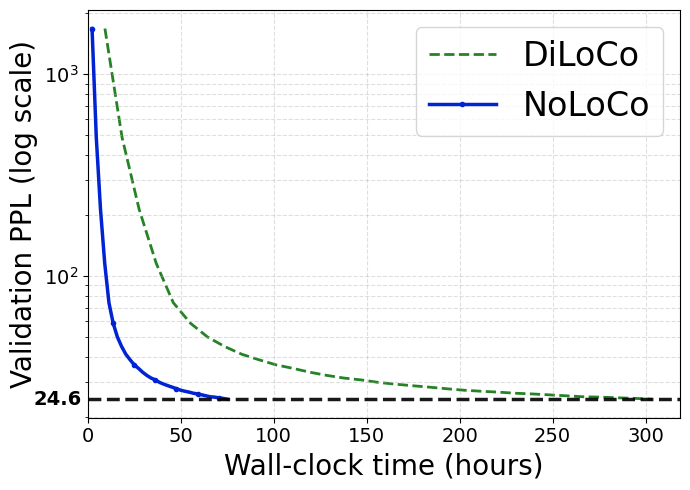}
        \caption{\hspace*{-1.1em}}
        \label{fig:wallclock_100mbps}
    \end{subfigure}
    \hfill
    \begin{subfigure}[t]{0.32\textwidth}
        \centering
        \includegraphics[width=\linewidth]{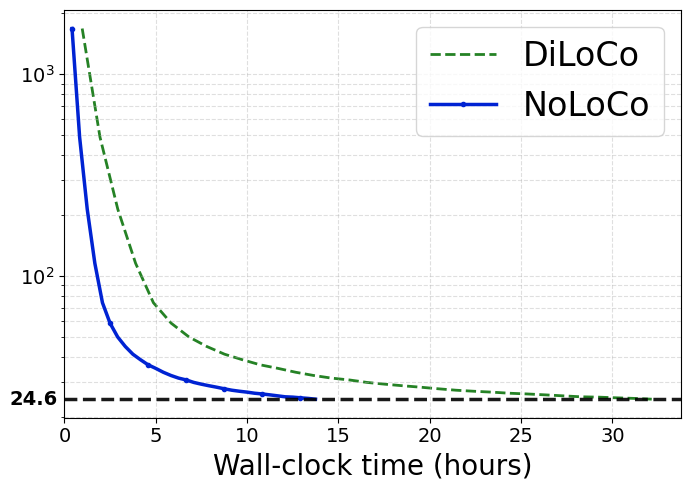}
        \caption{\hspace*{-0.8em}}
        \label{fig:wallclock_1gbps}
    \end{subfigure}
    \hfill
    \begin{subfigure}[t]{0.32\textwidth}
        \centering
        \includegraphics[width=\linewidth]{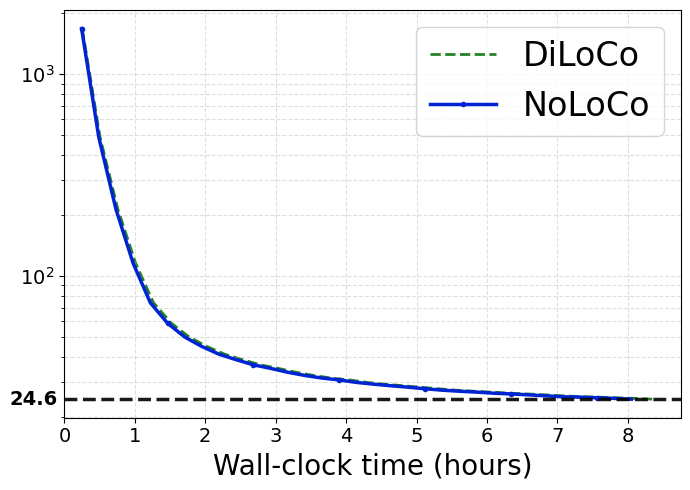}
        \caption{\hspace*{-0.8em}}
        \label{fig:wallclock_100gbps}
    \end{subfigure}
    \caption{Wall-clock convergence of DiLoCo and NoLoCo at three GPU interlink bandwidths (1.3B Llama model, C4 dataset, 32 workers, $H{=}100$). $(a)$ $100\,\mathrm{Mb/s}$; $(b)$ $1\,\mathrm{Gb/s}$; $(c)$  $100\,\mathrm{Gb/s}$.}
    \label{fig:wallclock}
\end{figure}

Fig.~\ref{fig:wallclock} makes this concrete by plotting validation PPL against wall-clock time for both methods at three representative bandwidths ($100\,\mathrm{Mb/s}$, $1\,\mathrm{Gb/s}$, and $100\,\mathrm{Gb/s}$). The PPL trajectories come from actual NoLoCo and DiLoCo training runs, with the time axes at $1\,\mathrm{Gb/s}$ and $100\,\mathrm{Mb/s}$ rescaled using the per-outer-step times in Table~\ref{tab:latency}. The dotted horizontal line marks the final converged PPL.

\paragraph{Theoretical and simulated speed-up for large number of workers.}
\label{sec:th_speedup}
We derive a theoretical speed-up for NoLoCo's local averaging relative to tree all-reduce. With $n$ workers and a per-message send time $t_c$, tree all-reduce takes $t_{\text{all}} \approx 2 t_c \log_2(n)$, while local averaging in pairs takes $2 t_c$ (yielding a ratio of $\log_2(n)$). Consistent with this, the NoLoCo/DiLoCo step-time ratio in Table~\ref{tab:latency} approaches $\log_2(32) = 5$ for the $100\,\mathrm{Mb/s}$ bandwidth case (note that as the network speed decreases, all-reduce dominates the step time).

To assess how this advantage scales, we simulate variable communication latency and asynchronous worker arrival for world sizes from $4$ up to $1024$ accelerators (see Appendix~\ref{appendix:latency} for details). Fig.~\ref{fig:latency:A} plots the ratio of expected tree-reduction time to expected local-averaging time, across different world sizes and message-send-time variances. We can see that NoLoCo's advantage over tree-reduction increases with world size, and increases further when communication times are more variable. Fig.~\ref{fig:latency:B} isolates the cost of global blocking communication: even ignoring all-reduce and local-averaging time, DiLoCo falls behind NoLoCo by $\sim 20\%$ at $1024$ accelerators with $100$ inner steps per outer step, because workers must wait for the slowest one at each outer step. Both effects compound, so the simulated gap between NoLoCo and DiLoCo widens substantially at the scale of training runs.

\begin{figure}[ht]
    \centering
    \begin{subfigure}[t]{0.495\linewidth}
        \centering
        \includegraphics[height=4.5cm]{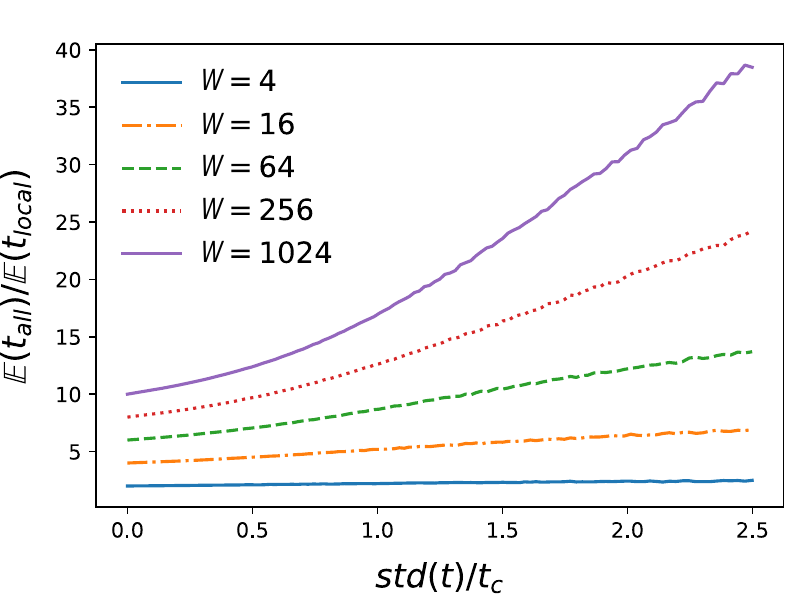}
        \caption{\hspace*{-0.8cm}}
        \label{fig:latency:A}
    \end{subfigure}
    \begin{subfigure}[t]{0.495\linewidth}
        \centering
        \includegraphics[height=4.5cm]{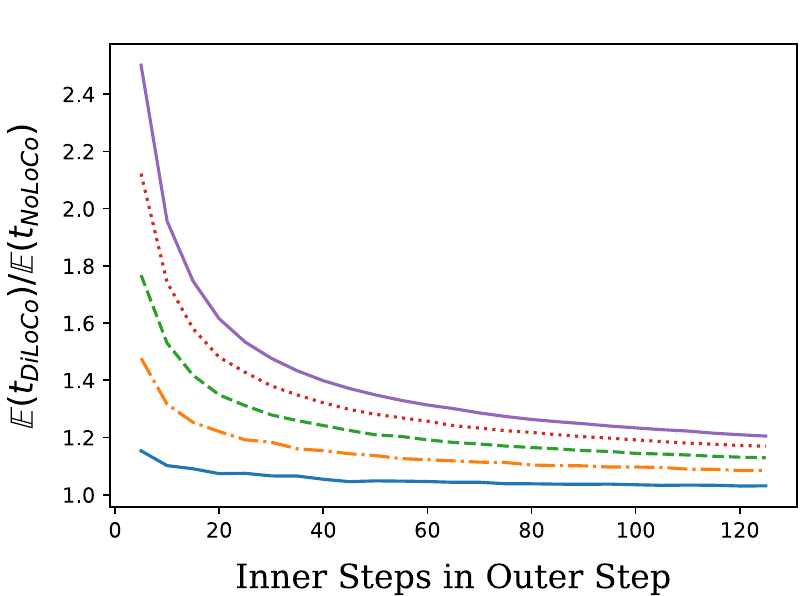}
        \caption{\hspace*{-0.8cm}}
        \label{fig:latency:B}
    \end{subfigure}
    \caption{
    $(a)$ Simulated ratio of expected tree-reduce to local-averaging time, where $W$ is the world size and $t_c = \exp(\mu + \sigma^2/2)$ the expected per-message time. $(b)$ Simulated ratio of total DiLoCo and NoLoCo training times, excluding communication overhead. All runs use $500$ outer steps, variable inner steps, and inner-step latency $\mathrm{LogNormal}(\mu = 1, \sigma^2 = 0.5)$. See Appendix~\ref{appendix:latency} for details.
    }
    \label{fig:appendix:latency}
\end{figure}

\section{Ablation Experiments}

\begin{figure}[t]
    \centering
    \makebox[\textwidth][c]{\hspace{-1em}
    \begin{subfigure}[t]{0.24\textwidth}
        \centering
        \includegraphics[width=1\linewidth]{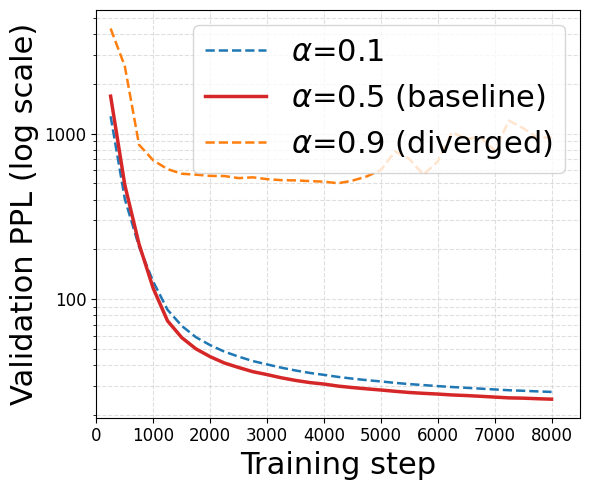}
        \caption{\hspace*{-0.8cm}}
        \label{fig:abl_alpha}
    \end{subfigure}\hspace{1em}%
    \begin{subfigure}[t]{0.24\textwidth}
        \centering
        \includegraphics[width=1\linewidth]{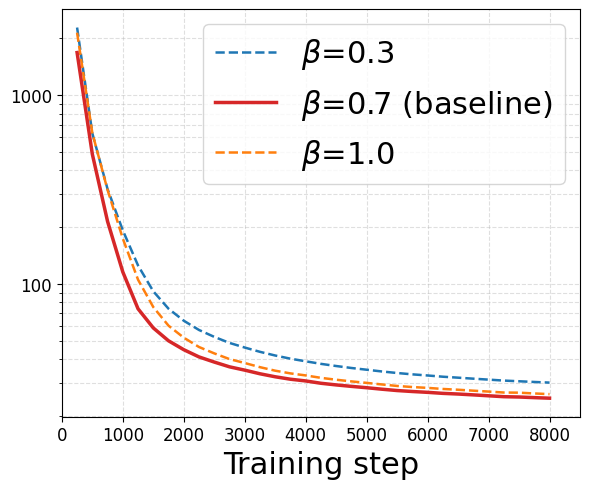}
        \caption{\hspace*{-0.7cm}}
        \label{fig:abl_beta}
    \end{subfigure}\hspace{1em}%
    \begin{subfigure}[t]{0.24\textwidth}
        \centering
        \includegraphics[width=1\linewidth]{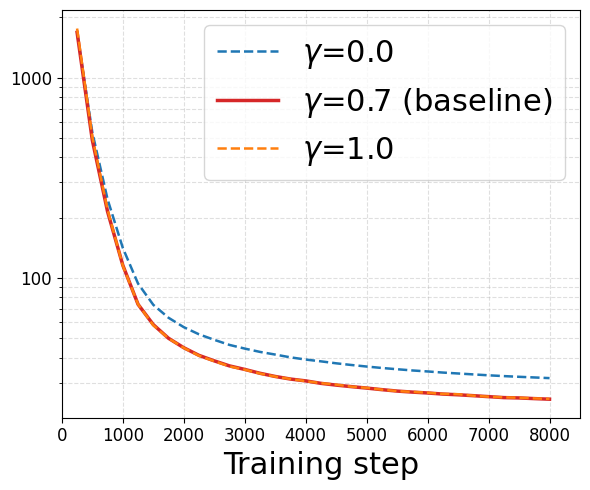}
        \caption{\hspace*{-0.7cm}}
        \label{fig:abl_gamma}
    \end{subfigure}\hspace{1em}%
    \begin{subfigure}[t]{0.24\textwidth}
        \centering
        \includegraphics[width=1\linewidth]{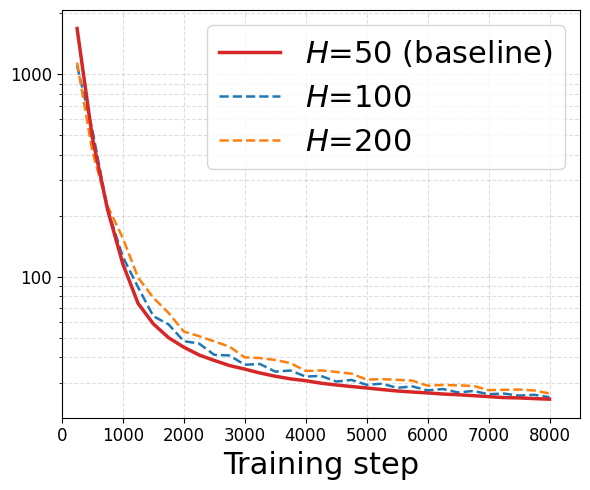}
        \caption{\hspace*{-0.7cm}}
        \label{fig:abl_H}
    \end{subfigure}%
    }
    \caption{NoLoCo hyperparameter ablations on a 1.3B model trained on C4 with DP=8, PP=2. $(a)$ Outer momentum $\alpha$; $(b)$ outer learning rate $\beta$; $(c)$ weight averaging $\gamma$; $(d)$ inner steps $H$. Each panel sweeps a single hyperparameter while holding the others at the baseline.}
    \label{fig:ablations}
\end{figure}

\paragraph{Hyperparameter ablations.}
\label{sec:hparam_ablations}
The PPL numbers in Section~\ref{sec:results} are subject to the training hyperparameters. To justify the hyperparameter choices and assess the sensitivity of our results to these choice, we performed additional hyperparameter sweeps on the local weight averaging parameter $\alpha$, the outer learning rate $\eta$, the outer momentum $\beta$, and the number of inner steps per outer step $H$. We use the medium (1.3B) model on the C4 dataset. The corresponding plots are in Fig.~\ref{fig:ablations}

\paragraph{Effects of local group size.}
\label{sec:n_ablation}

\begin{wrapfigure}{r}{0.5\textwidth}
\vspace{-1.1em}
    \centering
    \includegraphics[width=\linewidth]{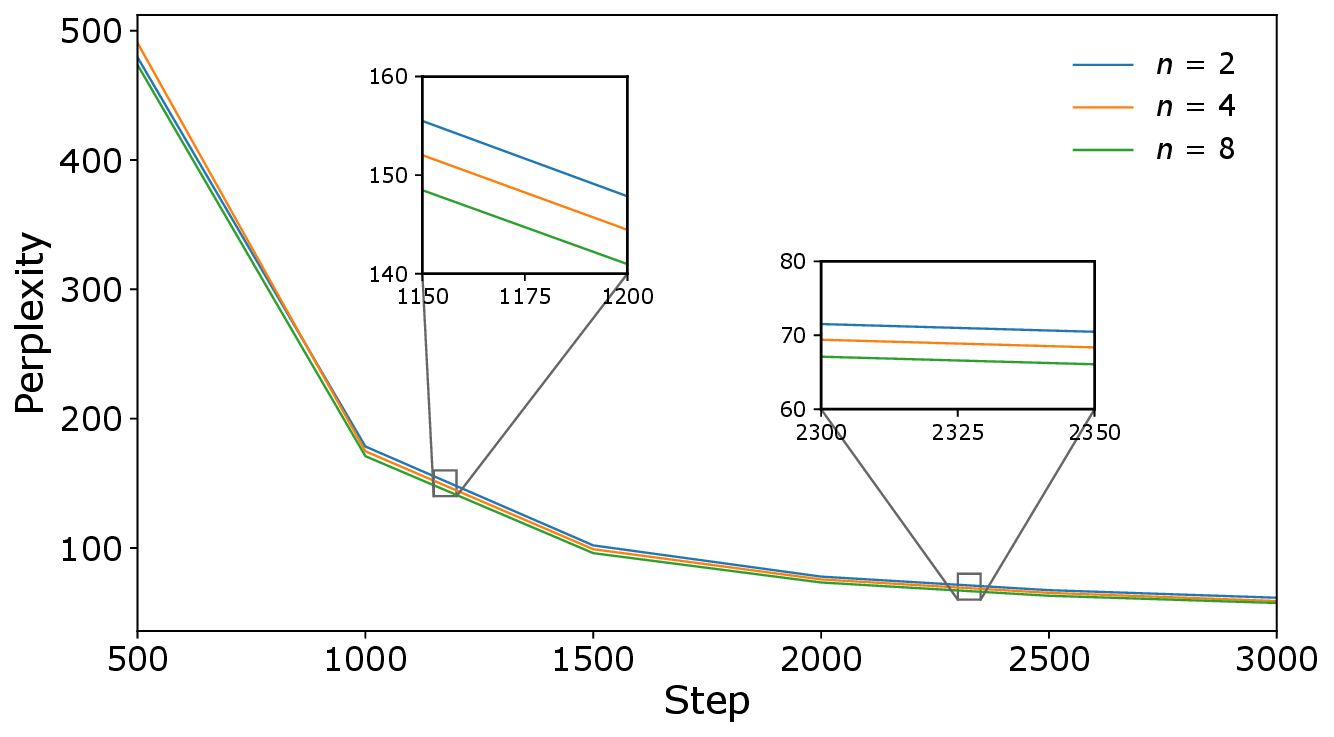}
    \caption{Perplexity of the 125M model for different local group sizes $n$. To eliminate the effects of random pipeline routing, we set the DP world size to match the number of accelerators (i.e.\ 8).}
    \label{fig:n_ablation}
\end{wrapfigure}

We conduct another ablation study to examine the effect of the local group size \(n\). Due to resource constraints, use the 125M model. To isolate the effect of \(n\), we set the data-parallel world size equal to the number of available accelerators, resulting in a single pipeline stage and effectively eliminating random pipeline routing. We train on 0.5M tokens from the English partition of the C4 training set and evaluate at every 500 steps on 1M tokens from the corresponding validation set. Fig.~\ref{fig:n_ablation} reports validation perplexities for \(n \in \{2,4,8\}\). For the model sizes and hyperparameter settings here, perplexity improves as $n$ grows, but the effect is weak. We expect this monotonic relationship to be more pronounced for larger models and datasets.

\paragraph{Effects of random pipeline routing.}
\label{sec:rand_pp_ablation}
To analyze the effect of random routing in the PP communication, we perform an ablation study comparing random routing with fixed routing. For fixed routing, nodes only send and receive values from fixed prior and subsequent model stages, as in standard setups \cite{huang2019gpipe,narayanan2021efficient,sun2024seq1f1b}. We also remove the outer optimizer synchronization (for both routing methods), meaning nodes in the same stage never exchange information directly. Thus, without random routing the setup is the same as running separate training jobs without DP. Using this setup, we repeat experiments for Reddit using the small (125M) and medium (1.3B) model sizes.

\begin{figure}[t]
    \centering
    \begin{subfigure}[t]{0.495\linewidth}
        \centering
        \includegraphics[height=4.3cm]{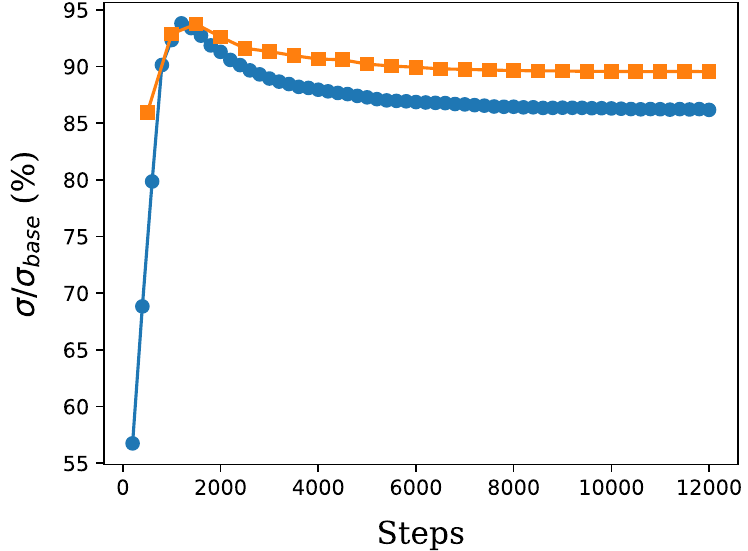}
        \caption{\hspace*{-0.9cm}}
        \label{fig:path_results:A}
    \end{subfigure}
    \begin{subfigure}[t]{0.495\linewidth}
        \centering
        \includegraphics[height=4.3cm]{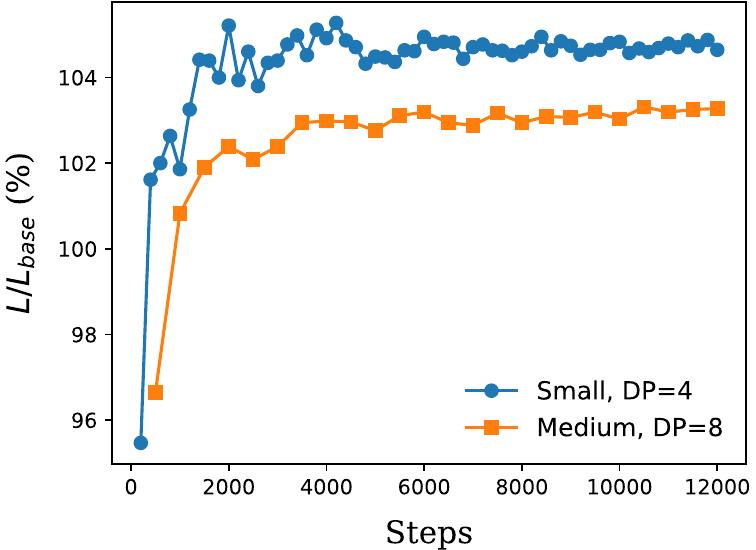}
        \caption{\hspace*{-1cm}}
        \label{fig:path_results:B}
    \end{subfigure}
    \caption{Results from training without any outer optimizer steps. The baseline loss and model replica variance is computed from DP number of independent runs. The combined training is only using the random pipeline routing. Figures show the ratio of $(a)$ weight standard deviations and $(b)$ validation perplexities  between the random and fixed routing.
    }
    \label{fig:appendix:path}
\end{figure}

We present the results in Fig. \ref{fig:appendix:path}. We observe that, for the small model, the standard deviation is $\sim 15\%$ lower than in the same run without random routing between different PP pipelines. This effect becomes less pronounced for the medium model with larger DP world size (namely $\sim 10\%$ lower standard deviation). Thus we observe that through PP, nodes in the same stage implicitly exchange information about their weights, without directly synchronizing. We attribute the faster convergence observed in Fig. \ref{fig:ppl_results} to this fact.

\section{Summary}

We proposed a novel low communication decentralized training method, NoLoCo, that requires only subgroup synchronization in outer optimizer steps and avoids collective all-to-all communication. 
While reducing the synchronisation group, to prevent divergence of model weights, we introduced a modification to the Nesterov optimizer used in the outer step. We provide a convergence proof of NoLoCo and show that its convergence rate is comparable with fully synchronized DP methods.

We compare NoLoCo with DiLoCo across model of $125$M, $1.3$B, and $6.8$B parameters, two language modeling datasets (C4 and Pushshift Reddit), and varying numbers of parallel workers.
We found that NoLoCo converges up to $4\%$ faster than DiLoCo in our experiments while not using any all-to-all communication. We hypothesize that this is because of the regularization effect coming from slight variations between different instances. 

The speed-up from removing the all-to-all communication outer optimizer steps depends on the standard deviation of the message sending latency and logarithmically on the DP worker count. We confirmed this behavior in our experiments where we observed $2\times$ and $4\times$ faster training step times with NoLoCo compared to DiLoCo using $1\,\mathrm{Gb/s}$ and $100\,\mathrm{Mb/s}$ interlink speeds, respectively.

We have demonstrated that local averaging is a viable option for training models with low bandwidth and high latency networks. 
Future work is needed to establish optimal hyper-parameters for NoLoCo and to explore how it can be combined with recent works on overlapping communication with compute or compression and quantisation techniques to further reduce the communication overhead.

\bibliography{bib2026} 
\bibliographystyle{plainnat}


\appendix

\section{Pseudo algorithm for random pipeline routing}
\label{app:pp_routing}

\begin{algorithm}[H]
\caption{Random pipeline routing (one micro-batch)}
\label{alg:random_routing}
\small
\begin{algorithmic}[1]
\State \textbf{// Forward pass}
\For{$\text{inputs} \in \text{batch}$}
    \State $r_{\text{recv}} \gets \textsc{RandomRank}(\text{dp\_rank} - 1, \text{seed})$
    \State $r_{\text{send}} \gets \textsc{RandomRank}(\text{dp\_rank} + 1, \text{seed})$
    \State $\text{inputs} \gets \textsc{MaybeRecvInputs}(r_{\text{recv}})$
    \State \textbf{with} \texttt{torch.no\_grad()}:
        \State \quad $\text{outputs} \gets \text{model}(\text{inputs})$
    \State \textsc{SaveForBackward}($\text{inputs}, r_{\text{recv}}, r_{\text{send}}$)
    \State \textsc{MaybeSendOutputs}($r_{\text{send}}$)
\EndFor
\Statex
\State \textbf{// Backward pass}
\For{$\_ \in \text{batch}$}
    \State $\text{inputs}, r_{\text{recv}}, r_{\text{send}} \gets \textsc{LoadForBackward}()$
    \State $\text{outputs} \gets \text{model}(\text{inputs})$
    \State $\text{grads} \gets \textsc{MaybeRecvGrads}(r_{\text{send}})$
    \State $\text{outputs}.\textsc{backward}(\text{grads})$
    \State \textsc{MaybeSendGrads}($\text{inputs}.\text{grad}, r_{\text{recv}}$)
\EndFor
\end{algorithmic}
\end{algorithm}

\section{Additional experiment details}
\label{Appx:hyper}

Details of the model's hyperparameters and their specific training hyperparameters can be found in Table \ref{table:model_parameters}. Global batch sizes and learning rates are taken from studies~\citet{shuai2024scaling,zhang2022opt}. Each run has a linear warm-up of $1000$ steps and a cosine learning rate schedule applied after the warm-up period to reduce the learning rate by one magnitude compared to the maximum learning rate. Unless otherwise stated, all training runs are done over $25,000$ optimizer steps. 

The total number of GPUs in each experiment (i.e.\ the world size) equals the product of the DP dimension (number of model replicas) and the PP dimension (number of pipeline stages). All experiments use Nvidia H100 clusters connected via NVLink. The bandwidths in the latency experiments reported in Section~\ref{section::latency} were achieved by throttling the network-interface-card using Cillium that achieves realistic low bandwidth network behavior.

\par The source code for running the experiments is available in \href{https://github.com/gensyn-ai/noloco}{GitHub}.
\begin{table}[h]
    \centering
        \caption{Summary of model hyper-parameters. Batch sizes are expressed in tokens.}
    \label{table:model_parameters}
    \begin{tabular}{l| r r r}
        Parameter Name & Small & Medium & Large \\
        \hline
        Hidden size & $768$ & $2048$ & $4096$ \\ 
        Layers & $12$ & $24$ & $32$ \\
        Intermediate size & $3072$ & $8192$ & $16,384$ \\
        Attention heads & $12$ & $32$ & $32$ \\
        Context length & $2048$ & $2048$ & $2048$ \\
        (Inner) Learning-rate & $0.0006$ & $0.0002$ & $0.00012$ \\
        Global batch size & $0.5 \mathrm{M}$ & $1 \mathrm{M}$ & $2 \mathrm{M}$ \\
        Transformer Parameters & $125\mathrm{M}$ & $1.3\mathrm{B}$ & $6.8\mathrm{B}$ \\
    \end{tabular}
\end{table}

\section{Additional comparisons to related work}
\label{appendix:lit_comparison}

We compare NoLoCo against several state-of-the-art gossip-like and asynchronous training methods, as well as optimizers using quantization and compression. The quantization and compression techniques complementary to NoLoCo and can be combined with it.

All comparisons train the medium (1.3B) model on C4 with $\text{DP}=8$ and $\text{PP}=2$. NoLoCo runs use the baseline hyperparameters from Section~\ref{sec:exp_setting}.

\paragraph{Epidemic Learning~\citep{de2023epidemic}.} 
Epidemic Learning~\citep{de2023epidemic} is a decentralized SGD method in which, at each round, every node sends its model update to a uniformly random sample of $n$ other nodes and averages the models it receives, yielding a fully randomized communication topology that changes each round.

Unlike EL, NoLoCo uses pairwise DiLoCo-style outer-step synchronization combined with random pipeline routing, hence exchanging less data, less often, and mixing replicas implicitly through PP rather than direct weight averaging.

\begin{figure}[h]
\centering
\includegraphics[height=0.3\linewidth]{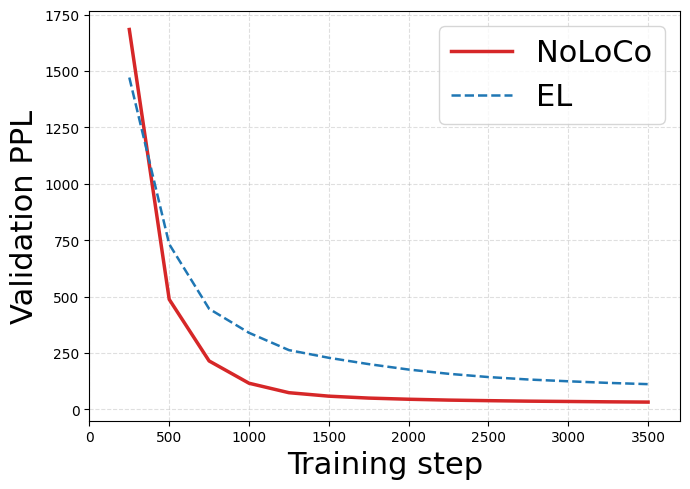}
\caption{Validation perplexity versus training step for NoLoCo and EL on a 1.3B-parameter model trained on C4. Both runs use the same setup: 16 GPUs partitioned as DP$=$8, PP$=$2, identical inner optimizer steps, learning-rate schedule, and per-worker batch size.} 
\label{fig:el_comparison}
\end{figure}

We implemented an EL baseline that performs pairwise weight averaging every $50$ steps, matching NoLoCo's communication budget. For the EL run, we set $\beta=0.7$ and the inner learning rate to $2\cdot 10^{-4}$. NoLoCo converges significantly faster (Fig.~\ref{fig:el_comparison}, which we attribute to NoLoCo's outer-gradient-based aggregation rather than direct weight averaging.

\paragraph{PALSGD~\citep{naganuma2025pseudoasync}.}
Both PALSGD and NoLoCo extend DiLoCo to reduce communication overhead. PALSGD uses a local proximal regularizer that pulls workers toward a stale global model copy, while NoLoCo employs momentum-level correction, random pipeline routing, and random subgroup synchronization. Because these operate on different aspects of training, they are in principle complementary and could be combined.

\begin{figure}[h]
\centering
\includegraphics[height=0.3\linewidth]{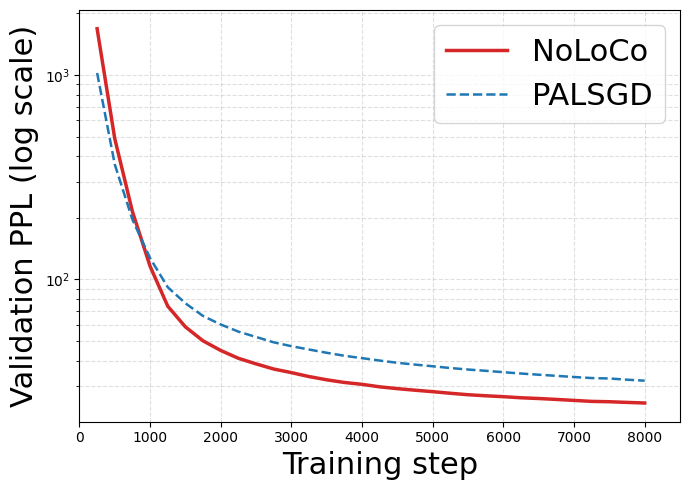}
\caption{Validation perplexity versus training step for NoLoCo and PALSGD on a 1.3B-parameter model trained on C4. Both runs use the same setup: 16 GPUs partitioned as DP$=$8, PP$=$2, identical inner optimizer steps, learning-rate schedule, and per-worker batch size.} 
\label{fig:palsgd_comparison}
\end{figure}

For \textsc{PALSGD} we use $H = 50$ inner steps per outer step, an inner learning rate of $2\times10^{-4}$, outer learning rate $\beta = 0.7$, outer momentum $\alpha = 0.9$, pseudo-synchronization probability $p = 0.05$, and pseudo-sync mixing rate $\eta = 0.1$. NoLoCo exceeds the convergence quality of PALSGD in our setup (Fig.~\ref{fig:palsgd_comparison})

For context, the original PALSGD experiments use $8$M and $125$M GPT-Neo models and report improvements only up to $H = 64$ inner steps, with degradation at larger values; DiLoCo, by contrast, reports results for $H$ ranging from $50$ to $2000$. Our results are therefore not contradictory with the PALSGD findings.

\paragraph{SparseLoCo~\citep{sarfi2025communication}.}
SparseLoCo reduces communication cost via top-$k$ gradient sparsification, $2$-bit quantization, and error-feedback–based local outer-momentum approximations. These optimisations cut the size of each outer message significantly relative to a bf16 DiLoCo step, but the underlying synchronisation primitive is unchanged: every outer step still consists of a global ring all-reduce in which all workers must finish their $H$ inner steps, exchange the (now-compressed) update, and barrier together before any worker can proceed. SparseLoCo therefore attacks the \emph{bandwidth} axis of  communication cost while leaving the \emph{coupling} axis untouched.

NoLoCo modifies the orthogonal axis. Each worker sends weights to $n-1$ random partners (with possibly $n=2$) and eliminates the issue of a global barrier: a transient bandwidth dip or a single straggler can no longer stall the entire cohort. Nonetheless, NoLoCo's per-byte volume is essentially the same as DiLoCo; the speed-up arises from removing synchronisation, not from sending fewer bits.   

Because SparseLoCo and NoLoCo modify disjoint aspects of communication (SparseLoCo modifies \emph{what} is communicated, NoLoCo modifies \emph{which workers communicate and when}) they are composable: a ``Sparse-NoLoCo'' that exchanges sparse, quantised, error-corrected deltas pairwise would compound both savings, and we view such a combination as a natural direction for future work.

\begin{figure}[h]
\centering
\includegraphics[height=0.3\linewidth]{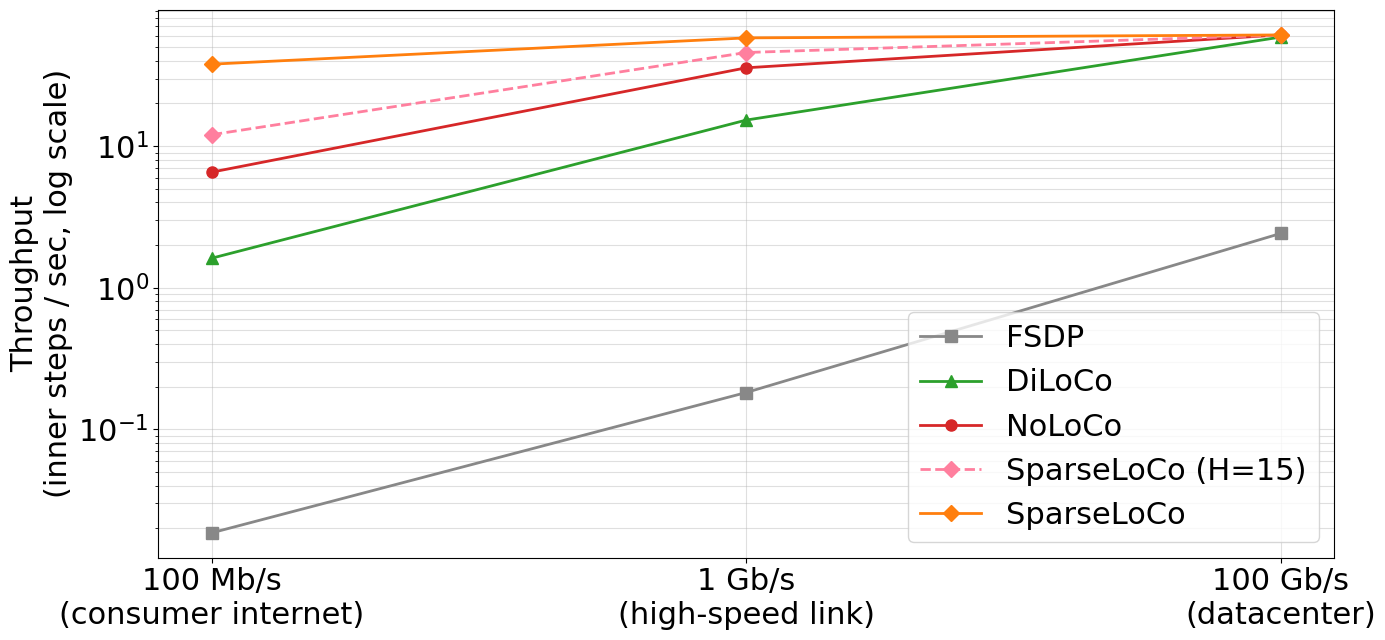}
\caption{Throughput (inner steps/sec, log scale) versus interlink bandwidth for FSDP, DiLoCo, NoLoCo, and SparseLoCo on a 1.3B model. FSDP, DiLoCo, and NoLoCo timings come from Table~\ref{tab:latency}; SparseLoCo is estimated by scaling DiLoCo's empirical communication cost by the SparseLoCo compression ratio ($\sim\!1.66\%$). At $100$\,Gb/s all local-SGD methods are compute-bound; as bandwidth drops, NoLoCo and SparseLoCo open a gap over DiLoCo via complementary mechanisms.} 
\label{fig:sparseloco_comparison}
\end{figure}

For the present comparison we simply report throughput in Fig.~\ref{fig:sparseloco_comparison}, where NoLoCo and SparseLoCo achieve comparable wall-clock gains over DiLoCo and FSDP via fundamentally different mechanisms. We report throughput (inner steps per second) as a function of network bandwidth, computed as $\text{Throughput} = H / T_{\text{outer}}(B)$. For FSDP, DiLoCo, and NoLoCo, we use the empirical per-step times from Table~\ref{tab:latency} directly. For SparseLoCo, we estimate throughput using the compression ratio from \citet[Table~2]{sarfi2025communication}, which gives a compression ratio of $1.66\%$. We scale DiLoCo communication time accordingly: \begin{equation*}
    T_{\text{SparseLoCo}}(B) = T_{\text{DiLoCo}}(B) \times 0.0166.
\end{equation*}
We additionally report SparseLoCo throughput at $H = 15$, matching the configuration of~\citet{sarfi2025communication}.

\paragraph{DN+DyLU (Async LocSGD)~\citep{liu2024asynclocalsgd}.}
DN+DyLU and NoLoCo address complementary bottlenecks: DN+DyLU targets the straggler effect, NoLoCo targets communication volume. Therefore, comparing them in either setting would disadvantage the other. 

The two approaches are in fact complementary for large-scale decentralized training, which must handle both variable worker speeds and prohibitive all-reduce costs. DN+DyLU exhibits diminishing returns as worker count grows, which the authors attribute to the sequential application of pseudo-gradients: with $N$ workers, each update is applied one at a time, and the compounding momentum increasingly distorts the parameter update as $N$ grows (Eq.~5 in~\citep{liu2024asynclocalsgd}). NoLoCo's communication pattern sidesteps this issue, since synchronization always involves a pair of workers regardless of $N$. For equivalent total communication cost, NoLoCo's pairwise syncs can also be performed at higher frequency than DiLoCo's all-reduce, reducing the maximum expected divergence. We hypothesize that this could mitigate the diminishing returns observed by~\citet{liu2024asynclocalsgd}.

\paragraph{GWTF~\citep{blagoev2025flow}.}
Go With The Flow (GWTF) offers efficient and churn-tolerant routing for decentralized pipeline-parallel training. We view this work as complementary to ours: GWTF's routing algorithm could replace our random routing during pipeline-parallel communication, potentially reducing per-iteration communication time.

\section{Convergence analysis}
\label{section::convergence_analsis}

\subsection{Expected value and variance of inner iteration}

The proposed method in this study aims to lower the communication overhead of data parallel training via the modified Nesterov momentum expression given by Eq. \ref{eq:delta}. This appendix shows proof that the modified version converges to the vicinity of the real optima when the hyperparameters are chosen appropriately. We structure the proof as follows: this section gives background context and derives expected value and variance for the inner iterations weights; Section \ref{appendix:sectA2} provides convergence proof for the expected value of slow weights; and finally \ref{appendix:sectA3} shows proof for the variance of the slow weights.

To show this we assume a stochastic loss from prior studies~\cite{schaul2013no,wu2018understanding,zhang2019lookahead}:
\begin{equation}
    \mathcal{L}(\theta) = \frac{1}{2}(\theta-c)^{\mathrm{T}} \mathrm{A} (\theta-c),
\end{equation} where $c\sim \mathcal{N}(0, \Sigma)$ is a vector valued random variable; $\mathrm{A}$ is a positive definite symmetric matrix; and the covariance matrix $\Sigma$ is a constant~\cite{zhang2019lookahead}. With the above loss function, the minimum value is obtained at $\theta=0$. Gradient of the loss function is given by:
\begin{equation}
    \nabla_{\theta} \mathcal{L}(\theta) = \mathrm{A} (\theta-c) \sim \mathcal{N}(\mathrm{A} \theta,  \mathrm{A}\Sigma \mathrm{A} ).
\end{equation}

For this analysis, we assume that the inner optimizer is using the stochastic gradient descent with a constant learning rate. Stochastic gradient descent updates the (fast) weights according to following rule:
\begin{equation}
    \theta^{(k+1)}_{t, i} =\left( \mathrm{I} - \omega \mathrm{A} \right) \theta^{(k)}_{t, i} + \omega \mathrm{A} c,
\end{equation} where $(k)$ denotes the (inner) iteration index and $\omega$ is the inner learning rate. With these assumptions, expected value and variance of the fast model weights after one inner iteration are~\cite{wu2018understanding}:
\begin{eqnarray}
\label{eq:appendix:it1}
\mathbb{E}(\theta_{t+1, i}^{(j+1)}) &=& \left( \mathrm{I} - \omega \mathrm{A} \right) \mathbb{E}(\theta_{t+1, i}^{(j)}), \\
\label{eq:appendix:it2}
\mathbb{V}(\theta_{t+1, i}^{(j+1)}) &=& \left( \mathrm{I} - \omega \mathrm{A} \right) \mathbb{V}(\theta_{t+1, i}^{(j)}) \left( \mathrm{I} - \omega \mathrm{A} \right) + \omega^2 \mathrm{A} \Sigma \mathrm{A},
\end{eqnarray} where $\mathrm{I}$ is the identity matrix, and $\theta_{t+1, i}^{(0)} = \phi_{t,i}$. To simplify the notation, we define following shorthands:
\begin{eqnarray}    
    \mathrm{B} &=& \mathrm{I} - \omega \mathrm{A}, \\
    \mathrm{U} &=& \omega^2 \mathrm{A} \Sigma \mathrm{A}.
\end{eqnarray} Solving Eqs. \ref{eq:appendix:it1} and \ref{eq:appendix:it2} for expected value and variance, we obtain:
\begin{eqnarray}
\label{eq:appendix:it1a}
\mathbb{E}(\theta_{t+1, i}^{(j+1)}) &=& \mathrm{B}^j \mathbb{E}(\theta_{t+1, i}^{(0)}), \\
\label{eq:appendix:it2a}
\mathbb{V}(\theta_{t+1, i}^{(j+1)}) &=& \mathrm{B}^{j} \mathbb{V}(\theta_{t+1, i}^{(0)})\mathrm{B}^{j} + \mathcal{F}^{-1}\left( \mathrm{U} - \mathrm{B}^{j}\mathrm{U}\mathrm{B}^{j} \right),
\end{eqnarray} where $\mathcal{F}$ is a linear function given by:
\begin{equation}
\mathcal{F}\left( \mathrm{X}\right) =  \mathrm{X} - \mathrm{B}\mathrm{X}\mathrm{B}.
\end{equation} $\mathcal{F}$ is invertible when $\omega > 0$ and $\mathcal{F}\left( \mathrm{X}\right) \equiv 0$ when $\omega=0$.

From Eqs. \ref{eq:appendix:it1a} and \ref{eq:appendix:it2a} we can solve for the outer gradient:

\begin{eqnarray}
    \mathbb{E}(\Delta_{t,i}) &=&  \mathbb{E}(\theta_{t+1,i}^{(m)} - \phi_{t,i}) \\
    &=& \left( \mathrm{I} - \omega \mathrm{A} \right)^m \mathbb{E}(\phi_{t,i}) - \mathbb{E}(\phi_{t,i}) \\
    &=& \left( \left( \mathrm{I} - \omega \mathrm{A} \right)^m - \mathrm{I} \right) \mathbb{E}(\phi_{t,i}),
    \label{eq:appendix:delta}
\end{eqnarray} where $m$ is the number of inner optimizer steps in one outer optimizer step. Similarly for variance,

\begin{eqnarray}
    \mathbb{V}(\Delta_{t,i}) &=& \mathbb{V}\left(\sum_{k=1}^{m} \theta_{t+1,i}^{(k)} - \theta_{t+1,i}^{(k-1)}\right) = \mathbb{V}(-\omega\sum_{k=0}^{m-1} \nabla_{\theta} \mathcal{L}(\theta_{t+1,i}^{(k)})) \\
    &\approx& \omega^2 \sum_{k=0}^{m-1} \mathbb{V}(\nabla_{\theta} \mathcal{L}(\theta_{t+1,i}^{(k)})) \\
    &=& \omega^2 \mathrm{A} \sum_{k=0}^{m-1} \left( \mathbb{V}(\theta_{t+1, i}^{(k)}) + \Sigma \right)  \mathrm{A} \\
    &=& \omega^2 \mathrm{A} \left( \sum_{k=0}^{m-1} \mathrm{B}^{k-1} \mathbb{V} (\theta_{t+1, i}^{(0)})  \mathrm{B}^{k-1} + \mathcal{F}^{-1}\left( \mathrm{U} - \mathrm{B}^{k-1}\mathrm{U}\mathrm{B}^{k-1} \right) \right) \mathrm{A} + m \mathrm{U} \\
    &=& \omega^2 \mathrm{A} \left( \sum_{k=0}^{m-1} \mathrm{B}^{k}  \mathbb{V}(\theta_{t+1, i}^{(0)}) \mathrm{B}^{k} \right) \mathrm{A} + \mathrm{R}' \\
    &=& \omega^2  \mathrm{A} \mathcal{F}^{-1}\left(\mathbb{V}(\theta_{t+1, i}^{(0)}) - \mathrm{B}^{k}  \mathbb{V}(\theta_{t+1, i}^{(0)}) \mathrm{B}^{k} \right) \mathrm{A} + \mathrm{R}', 
    \label{eq:appendix:delta_var}
\end{eqnarray} where $\mathrm{R}'$ is a constant matrix depending on $\omega$, $\Sigma$, $\mathrm{A}$, and $m$. In Eq. \ref{eq:appendix:delta_var} we neglected covariance of non-consecutive fast weights similar to study~\cite{zhang2019lookahead}. Eq. \ref{eq:appendix:delta_var} can be simplified by writing the variance matrix as a vector using following notation:
\begin{equation}
    \mathbb{U}(\theta) \equiv \mathrm{vec}(\mathbb{V}(\theta)) = \lbrack \mathbb{V}(\theta)_1, \mathbb{V}(\theta)_2, \cdots \rbrack,
\end{equation} where $\mathbb{V}(\theta)_1$ and $\mathbb{V}(\theta)_2$ are column vectors of the covariance matrix $\mathbb{V}(\theta)$; and $\mathrm{vec}(\cdot)$ denotes converting a matrix to a vector by concatenating all the column vectors. With this notation we obtain:
\begin{eqnarray}
    \mathbb{U}(\Delta_{t,i}) &=& \mathrm{vec} \left( \omega^2  \mathrm{A} \mathcal{F}^{-1}\left(\mathbb{V}(\theta_{t+1, i}^{(0)}) - \mathrm{B}^{k}  \mathbb{V}(\theta_{t+1, i}^{(0)}) \mathrm{B}^{k} \right) \mathrm{A} + \mathrm{R}'\right) \\
    &=& \omega^2 \mathrm{A}\otimes \mathrm{A} \left( \mathrm{I} - \mathrm{B} \otimes \mathrm{B} \right)^{-1} \left( \mathrm{I} - \mathrm{B}^{k} \otimes \mathrm{B}^{k}\right) \mathbb{U}(\theta_{t+1, i}^{(0)}) + \mathrm{vec}(\mathrm{R}'),
\end{eqnarray} where $\otimes$ denotes Kronecker product. We also define:
\begin{equation}
    \mathrm{B}_{\mathbb{V}} = \omega^2 \mathrm{A}\otimes \mathrm{A} \left( \mathrm{I} - \mathrm{B} \otimes \mathrm{B} \right)^{-1} \left( \mathrm{I} - \mathrm{B}^{k} \otimes \mathrm{B}^{k}\right).
\end{equation}

\subsection{Expected value of outer iteration}
\label{appendix:sectA2}

We proceed to show following:
\begin{theorem}
    As the outer iteration count $t$ tends to infinity, the expected value of the slow weights $\mathbb{E}(\phi_{t,i}) \to 0$.
\end{theorem} 

We remind the reader that the modified Nesterov momentum used in this study is given by:
\begin{equation}
    \delta_{t,i} = \alpha \delta_{t-1,i} - \frac{\beta}{n} \left( \sum_j {\Delta_{t,j}} \right) - \gamma \left(\phi_{t,i} - \frac{1}{n} \sum_j \phi_{t,j} \right),
    \label{eq:appendix:momentum}
\end{equation} where $n$ is the group size used for the sample means. 

The different realizations of the slow weights, $\phi_{t,j}$, are generally not independent due to the path decomposition mechanism and due to the previous outer iterations. However, we will assume that they would be independent for convergence analysis, which introduces an error in the expected value and variance expressions. This error will become smaller when the data parallel world size becomes larger as the coupling between slow weights becomes weaker. 

The optimization method will start from the same instance of slow model weights; hence $\phi_{0,i} \equiv \phi_0$ as the first starting point. All instances of slow weights are updated by the same algorithm and hence the value of $\phi_{t,i}$ depends on two things: (1) what data is used to compute the inner gradients; and (2) what other instances are used in the sample averages. Both of these processes are identically random regarding different instances, and hence we assume that the slow weights have identical distributions. This has following corollaries.

\begin{lemma}
    The expected values of the slow weights $\phi_{t,i}$ satisfy: \( \mathbb{E}(\phi_{t,i} - \frac{1}{n} \sum_j \phi_{t,j}) = 0 \).
    \label{eq:appendix:expected_value1}
\end{lemma} We present a formal proof Lemma \ref{eq:appendix:expected_value1} in Appendix \ref{section:appendix:proof}. For the variances situation is more complex, and we assume following conjecture based on the above informal reasoning:
\begin{conjecture}
        The variances of the slow weights \(\phi_{t,i}\) satisfy: \( \mathbb{V}(\phi_{t,i} - \frac{1}{n} \sum_j \phi_{t,j}) \approx 2 \left(\frac{n-1}{n}\right)^2\mathbb{V}(\phi_{t,i}) \).
\end{conjecture}

The outer gradients are fundamentally dependent on the current slow weights. We neglect the dependency of the instances outer gradients on other gradients slow weights as before, which will introduce another error in the approximation. This error should diminish as the data parallel world size becomes larger and the coupling between different instances of slow weights becomes weaker. Similar to reasoning with the slow weights, we assume that the outer gradients have identical distributions. With these assumptions, the expected value of $\delta_{t,i}$ becomes:
\begin{eqnarray}
    \mathbb{E}(\delta_{t,i}) &=& \alpha \mathbb{E}(\delta_{t-1,i}) + \beta \mathbb{E}(\Delta_{t,i}), \\
    &=& \beta \sum_{k=0}^{t} \alpha^{t-k} \mathbb{E}(\Delta_{k,i}) \\
    &=& \beta \sum_{k=0}^{t} \alpha^{t-k}  \left( \mathrm{B}^m - \mathrm{I} \right) \mathbb{E}(\phi_{k,i})
\end{eqnarray} that is the same expression as for regular Nesterov momentum. Expected value of the next iteration slow weights becomes:

\begin{eqnarray}
    \mathbb{E}(\phi_{t+1,i}) &=& \mathbb{E}(\phi_{t,i}) + \mathbb{E}(\delta_{t,i}) \\
    &=& \mathbb{E}(\phi_{t,i}) + \beta \sum_{k=0}^{t} \alpha^{t-k}  \left( \mathrm{B}^m - \mathrm{I} \right) \mathbb{E}(\phi_{k,i}) \\
    &=& \mathbb{E}(\phi_{t,i}) + \beta \alpha^t \left( \mathrm{B}^m - \mathrm{I} \right) \sum_{k=0}^{t} \alpha^{-k} \mathbb{E}(\phi_{k,i}) \\
    &=& \mathbb{E}(\phi_{t,i}) + \beta \alpha^t \left( \mathrm{B}^m - \mathrm{I} \right) \left( \alpha^{-t}\mathbb{E}(\phi_{t,i}) + \sum_{k=0}^{t-1} \alpha^{-k} \mathbb{E}(\phi_{k,i}) \right) \\
    &=& \left( \mathrm{I} + \beta (\mathrm{B}^m - \mathrm{I}) \right) \mathbb{E}(\phi_{t,i}) \\
    && + \alpha \left( -\mathbb{E}(\phi_{t-1,i}) + \mathbb{E}(\phi_{t-1,i}) + \beta \alpha^{t-1} \left( \mathrm{B}^m - \mathrm{I} \right) \sum_{k=0}^{t-1} \alpha^{-k} \mathbb{E}(\phi_{k,i})  \right) \\
    &=& \left( \mathrm{I} + \beta (\mathrm{B}^m - \mathrm{I}) \right) \mathbb{E}(\phi_{t,i}) + \alpha \left( \mathbb{E}(\phi_{t,i}) - \mathbb{E}(\phi_{t-1,i}) \right)\\
    &=& \left( \mathrm{I} + \alpha \mathrm{I} + \beta (\mathrm{B}^m - \mathrm{I}) \right) \mathbb{E}(\phi_{t,i}) - \alpha \mathbb{E}(\phi_{t-1,i}) \\
    &=& \mathrm{D} \mathbb{E}(\phi_{t,i}) - \alpha \mathbb{E}(\phi_{t-1,i}),
\end{eqnarray}  where $\mathrm{D}=\mathrm{I} + \alpha \mathrm{I} + \beta (\mathrm{B}^m - \mathrm{I})$. Solving $\mathbb{E}(\phi_{t,i})$ from the above recursive equation using method of characteristics yields:
\begin{equation}
    \mathbb{E}(\phi_{t,i}) = \mathrm{C_1} r_1^t + \mathrm{C}_2r_2^t,
\end{equation} where $\mathrm{C}_1$ and $\mathrm{C}_2$ are constants independent of $t$; $r_1$ and $r_2$ are the matrix roots of the characteristic polynomial given by:
\begin{eqnarray}
    r_1 &=& \frac{1}{2} \left( \mathrm{D} + \sqrt{\mathrm{D}^2 - 4\alpha \mathrm{I}} \right), \\ 
    r_2 &=& \frac{1}{2} \left( \mathrm{D} - \sqrt{\mathrm{D}^2 - 4\alpha \mathrm{I}} \right).
\end{eqnarray} We note that $0< r_2 \le r_1 \le \mathrm{D}$ when $0 \le \alpha < 1$, hence it is sufficient to show that $\mathrm{D}^{t} \rightarrow 0$ that happens if and only if all eigen values of $\mathrm{D}$, $\mathcal{D}_i$, have less than unit absolute value. Recall, that $\mathrm{A}$ is symmetric and positive definite and hence has eigen value decomposition: $\mathrm{A} = \mathrm{Q} \Lambda \mathrm{Q}^T$ where $\mathrm{Q}$ is an orthogonal matrix and $\Lambda$ is a diagonal matrix with positive non-zero elements at diagonal. This yields:
\begin{eqnarray}
    \mathrm{B}^m &=& \left( \mathrm{I} - \omega \mathrm{Q} \Lambda \mathrm{Q}^T \right)^m  \\
        &=& \left( \mathrm{Q} (\mathrm{I} - \omega \Lambda) \mathrm{Q}^\mathrm{T} \right)^m \\
        &=& \mathrm{Q} \left(\mathrm{I} - \omega \Lambda\right)^m \mathrm{Q}^\mathrm{T}.
\label{eq:appendix:bm}
\end{eqnarray} Substituting Eq. \ref{eq:appendix:bm} to definition of $\mathrm{D}$ gives:
\begin{eqnarray}
    \mathrm{D} &=& \mathrm{I} + \alpha \mathrm{I} + \beta (\mathrm{B}^m - \mathrm{I}) \\
    &=& \mathrm{I} + \alpha \mathrm{I} + \beta \mathrm{Q}\left( \left(\mathrm{I} - \omega \Lambda\right)^m - \mathrm{I} \right)\mathrm{Q}^\mathrm{T} \\
    &=& \mathrm{Q}\left( \mathrm{I} + (\alpha - \beta) \mathrm{I} + \beta \left(\mathrm{I} - \omega \Lambda\right)^m \right) \mathrm{Q}^\mathrm{T},
\end{eqnarray} where we can identify that the eigen values are:
\begin{equation}
    \mathcal{D}_i = 1 + \alpha - (1 - (1 - \omega \Lambda_i)^m)\beta,
\end{equation} where $\Lambda_i > 0$ is the $i^{\mathrm{th}}$ eigen value of $\mathrm{A}$. Convergence of the expected value depends on the hyper-parameters $\alpha$, $\beta$, $\omega$, and $m$. When $m$ is sufficiently large and inner learning rate is chosen to satisfy $0 < \omega \Lambda_i \le 1$, sufficient condition is $\beta > \alpha$ and $\mathbb{E}(\phi_{t,i}) \rightarrow 0$ when $t \rightarrow \infty$. Thus the method converges to optimal solution.

\subsection{Variance of outer iteration}
\label{appendix:sectA3}

Finally, we will show following:
\begin{theorem}
    As the outer iteration count $t$ tends to infinity, the variance of the slow weights satisfy \(\mathbb{V}(\phi_{t,i}) \propto \omega^2\).
\end{theorem} 

The variance of slow weights is given by:
\begin{equation}
\mathbb{V}(\phi_{t+1,i}) = \mathbb{V}(\phi_{t,i}) + \mathbb{V}(\delta_{t,i}) + 2 \mathrm{Cov}(\phi_{t,i}, \delta_{t,i})
\label{eq:appendix:var}
\end{equation} We only consider direct dependencies of slow weights $\phi_{t,i}$ for the covariance term that becomes:
\begin{equation}
    \mathrm{Cov}(\phi_{t,i}, \delta_{t,i}) \approx -\gamma^2 \frac{n-1}{n} \mathbb{V}(\phi_{t,i}),
    \label{eq:appendix:delta_covariance}
\end{equation} where we omitted the covariance between the outer gradients and corresponding slow weights. The variance of the momentum term becomes:
\begin{eqnarray}
    \mathbb{U}(\delta_{t,i}) &\approx& \alpha^2 \mathbb{U}(\delta_{t-1,i}) + \frac{\beta^2}{n} \mathbb{U} (\Delta_{t,i}) + 2 \gamma^2 \left(\frac{n-1}{n}\right)^2\mathbb{U}(\phi_{t,i}) \\
    &=&\alpha^2 \mathbb{U}(\delta_{t-1,i}) + \left(\frac{\beta^2 \mathrm{B}_{\mathbb{V}}}{n} + 2 \gamma^2 \left(\frac{n-1}{n}\right)^2   \right) \mathbb{U}(\phi_{t,i}) + \frac{\beta^2}{n}\mathrm{vec}(\mathrm{R}') \\
    &=& \sum_{k=0}^{t-1} \alpha^{2(t-1-k)} \left(\left(\frac{\beta^2 \mathrm{B}_{\mathbb{V}}}{n} + 2 \gamma^2 \left(\frac{n-1}{n}\right)^2   \right) \mathbb{U}(\phi_{k,i}) + \frac{\beta^2}{n}\mathrm{vec}(\mathrm{R}')\right) \\
    &=& \alpha^{2t-2}\sum_{k=0}^{t-1} \alpha^{-2k} ( \mathrm{C}_{\mathbb{V}} \mathbb{U}(\phi_{k,i}) +  \mathrm{R}''),
    \label{eq:appendix:delta_variance}
\end{eqnarray} where we assumed that initial momentum is $\mathbb{V}(\delta_{0,i}) \equiv 0$, and  used shorthands:
\begin{eqnarray}
    \mathrm{C}_{\mathbb{V}} &=& \frac{\beta^2 \mathrm{B}_{\mathbb{V}}}{n} + 2 \gamma^2\left(\frac{n-1}{n}\right)^2,\\
    \mathrm{R}'' &=& \frac{\beta^2}{n}\mathrm{vec}(\mathrm{R}').
\end{eqnarray}

Substituting Eqs. \ref{eq:appendix:delta_covariance} and \ref{eq:appendix:delta_variance} into Eq. \ref{eq:appendix:var} yields:
\begin{eqnarray}
\mathbb{U}(\phi_{t+1,i}) &=& \mathbb{U}(\phi_{t,i}) -2 \gamma^2 \frac{n-1}{n} \mathbb{U}(\phi_{t,i}) +\alpha^{2t-2}\sum_{k=0}^{t-1} \alpha^{-2k} ( \mathrm{C}_{\mathbb{V}} \mathbb{U}(\phi_{k,i}) +  \mathrm{R}'') \\
 &=&  \left(1 -2 \gamma^2 \frac{n-1}{n} \right) \mathbb{U}(\phi_{t,i})
+\alpha^{2t-2}\sum_{k=0}^{t-1} \alpha^{-2k} ( \mathrm{C}_{\mathbb{V}} \mathbb{U}(\phi_{k,i}) +  \mathrm{R}'').
\label{eq:appendix:eqi}
\end{eqnarray} The last sum-term can be rearranged as follows:
\begin{eqnarray}
   && \alpha^{2t-2}\sum_{k=0}^{t-1} \alpha^{-2k} (\mathrm{C}_{\mathbb{V}} \mathbb{U}(\phi_{k,i}) +  \mathrm{R}'')  \\ 
    && = \mathrm{C}_{\mathbb{V}} \mathbb{U}(\phi_{t-1,i}) +  \mathrm{R}'' + \alpha^2 \left( \alpha^{2(t-2)} \sum_{k=0}^{t-2} \alpha^{-2k} (\mathrm{C}_{\mathbb{V}} \mathbb{U}(\phi_{k,i}) +  \mathrm{R}'') \right) \\
      && = \mathrm{C}_{\mathbb{V}} \mathbb{U}(\phi_{t-1,i}) +  \mathrm{R}'' + \alpha^2 \left(\mathbb{U}(\phi_{t,i}) - \left(1 -2 \gamma^2 \frac{n-1}{n} \right) \mathbb{U}(\phi_{t-1,i}) \right)\\
      &&=\alpha^2 \mathbb{U}(\phi_{t,i}) +  \left(\mathrm{C}_{\mathbb{V}} - \alpha^2 \left(1 -2 \gamma^2 \frac{n-1}{n} \right) \mathrm{I}\right) \mathbb{U}(\phi_{t-1,i}) + \mathrm{R}''.
      \label{eq:appendix:eqii}
\end{eqnarray} Substituting Eq. \ref{eq:appendix:eqii} to Eq. \ref{eq:appendix:eqi} gives:
\begin{equation}
    \mathbb{U}(\phi_{t+1,i}) = d_{\mathbb{V}} \mathbb{U}(\phi_{t,i}) +  \mathrm{E}_{\mathbb{V}} \mathbb{U}(\phi_{t-1,i}) + \mathrm{R}'',
\end{equation} where $d_{\mathbb{V}}$ and $\mathrm{E}_{\mathbb{U}}$ are given by:
\begin{eqnarray}
    d_{\mathbb{V}} &=& 1 + \alpha^2 -2 \gamma^2 \frac{n-1}{n}, \\
    \mathrm{E}_{\mathbb{V}} &=& \mathrm{C}_{\mathbb{V}} - \alpha^2 \left(1 -2 \gamma^2 \frac{n-1}{n}\right) \mathrm{I}.
    \label{eq:appendix:var_rec}
\end{eqnarray} Solving $\mathbb{U}(\phi_{t,i})$ by method of characteristics gives following solution:
\begin{equation}
    \mathbb{U}(\phi_{t,i}) = \mathrm{C}_1' v_1^t + \mathrm{C}_2' v_2^t + \mathrm{R}'',
\end{equation} where $\mathrm{C}_1'$ and $\mathrm{C}_2'$ are constants independent of $t$; $v_1$ and $v_2$ are the matrix roots of the characteristic polynomial given by:
\begin{eqnarray}
    v_1 &=& \frac{1}{2} \left( d_{\mathbb{V}}\mathrm{I} + \sqrt{d_{\mathbb{V}}^2\mathrm{I} - 4\mathrm{E}_{\mathbb{V}} } \right), \\ 
    v_2 &=& \frac{1}{2} \left( d_{\mathbb{V}}\mathrm{I} - \sqrt{d_{\mathbb{V}}^2\mathrm{I} - 4\mathrm{E}_{\mathbb{V}}} \right).
\end{eqnarray} 

For real roots, we have following bounds: $0 \leq \| v_2 \|_2 \leq \| v_1 \|_2 \leq |d_{\mathbb{V}}|$. For the variance to remain bounded as $t \rightarrow \infty$ we must have $|d_{\mathbb{V}}| < 1$. Solving for $\gamma$ we arrive at condition:
\begin{equation}
    \sqrt{\frac{n}{2(n-1)}} \alpha < \gamma < \sqrt{\frac{n}{2(n-1)}(2 +\alpha^2)}.
\label{eq:appendix:condition}
\end{equation} When hyper-parameters satisfy Eq. \ref{eq:appendix:condition}, $\mathbb{U}(\phi_{t,i}) \rightarrow \mathrm{R}''$ when $t \rightarrow \infty$. The leading order term of $\mathrm{R}''$ in terms of inner learning rate is $\propto \omega^2$. Hence, the variance $\|\mathbb{U}(\phi_{t,i})\|_2  \propto \omega^2$ when $\omega \rightarrow 0$ which proofs that the method converges to the correct optima and the variance of the optima is proportional to square of inner learning rate.

\section{Proof for identical model expected values}
\label{section:appendix:proof}
\par Reminder that after \(k+1\) steps, the expected fast weights are:
\begin{equation}
\mathbb{E}[\theta^{k+1}_{t+1,i}] = \mathrm{B}^k \mathbb{E}[\phi_{t,i}],
\end{equation} where $\mathrm{B} = \mathrm{I} - \omega \mathrm{A}$. We prove that for any arbitrary step \(t\) the expected difference between an arbitrary \(\phi_{t,i}\) and the average of several other \(\frac{1}{n}\sum^{n}_j\phi_{t,j}\) is 0:
\begin{equation}
\mathbb{E}[\phi_{t,i} - \frac{1}{n}\sum^{n}_j\phi_{t,j}] = \mathbb{E}[\phi_{t,i}] - \frac{1}{n}\sum_j^n\mathbb{E}[\phi_{t,j}] = 0.
\end{equation}

\par Reminder about some of the variables:
\begin{eqnarray}
\Delta_{t,i} &=& \theta^{k+1}_{t+1,i} -\phi_{t,i} \\
\phi_{t+1,i} &=& \phi_{t,i} + \delta_{t,i}\\
\delta_{t,i} &=& \alpha\delta_{t-1,i} - \frac{\beta}{n}\sum^{n}_j \Delta_{t,j} - \gamma\left(\phi_{t,i} - \frac{1}{n}\sum^n_j\phi_{t,j} \right).
\end{eqnarray}

\par It is trivial to see that at step 1 the property holds:
\begin{eqnarray}
\phi_{0,i} &=& \phi_{0} \\
\delta_{0,i} &=& 0 \\
\delta_{1,i} &=& - \frac{\beta}{n}\sum_j^n\left(\theta^{k+1}_{t+1,i} - \phi_0 \right) 
\mathbb{E}[\delta_{1,i}] \\ 
&=& - \frac{\beta}{n}\sum_j^n\left(\mathrm{B}^k\mathbb{E}[\phi_0] - \mathbb{E}[\phi_0] \right)\\ &=& -\beta(\mathrm{B}^k - \mathrm{I})\mathbb{E}[\phi_0], \text{ and } \\
\mathbb{E}[\phi_{1,i}] &=& \mathbb{E}[\phi_{0}] -\beta(\mathrm{B}^k - \mathrm{I})\mathbb{E}[\phi_0].
\end{eqnarray}
\par Since the value does not depend on \(i\), then the value will be identical across all replicas, thus \(\mathbb{E}[\phi_{1,i}] - \frac{1}{n}\sum_j^n\mathbb{E}[\phi_{1,j}] = 0\).

\par We proceed next to show that if the property holds for an arbitrary step \(T\) and for all steps prior to that \(t \in [0,T]\), then it also holds for \(T+1\).

\par At step \(T+1\) we have:
\begin{equation}
\delta_{T+1,i} = \alpha\delta_{t,i} - \frac{\beta}{n}\sum^{n}_j \Delta_{T,j} - \gamma\left(\phi_{T,i} - \frac{1}{n}\sum^n_j\phi_{T,j} \right).
\label{eq:appendix_b:deta_t_1}
\end{equation}

\par Taking expected value of Eq. \ref{eq:appendix_b:deta_t_1} yield:
\begin{equation}
\mathbb{E}[\delta_{T+1,i}] = \alpha\mathbb{E}[\delta_{T,i}] - \frac{\beta}{n}\sum^{n}_j \mathbb{E}[\Delta_{T,j}] - \gamma\left(\mathbb{E}[\phi_{T,i}] - \frac{1}{n}\sum^n_j\mathbb{E}[\phi_{T,j}] \right).
\end{equation}

\par The right term reduces to $0$ and thus we are left with:
\begin{eqnarray}
\mathbb{E}[\delta_{T+1,i}] &=& \alpha\mathbb{E}[\delta_{T,i}] - \frac{\beta}{n}(\mathrm{B}^k - \mathrm{I})\sum^{n}_j\mathbb{E}[\phi_T] \\
&=& \alpha\mathbb{E}[\delta_{T,i}] - \beta(\mathrm{B}^k - \mathrm{I})\mathbb{E}[\phi_T],
\end{eqnarray} and
\begin{equation}
   \mathbb{E}[\phi_{T+1,i}] = \mathbb{E}[\phi_{T,i}] + \alpha\mathbb{E}[\delta_{t,i}] - \beta(\mathrm{B}^k - \mathrm{I})\mathbb{E}[\phi_T].
\end{equation} Finally,
\begin{eqnarray}
\mathbb{E}[\phi_{T+1,i}] - \frac{1}{n}\sum_j^n\mathbb{E}[\phi_{T+1,j}] &=& 
\mathbb{E}[\phi_{T,i}] + \alpha\mathbb{E}[\delta_{T,i}] - \beta(\mathrm{B}^k - \mathrm{I})\mathbb{E}[\phi_T] \\ 
&&-  \frac{1}{n}\sum_j^n\left(\mathbb{E}[\phi_{T,j}] + \alpha\mathbb{E}[\delta_{T,j}] - \beta(\mathrm{B}^k - \mathrm{I})\mathbb{E}[\phi_T]\right) \\
&=&\alpha\mathbb{E}[\delta_{T,i}] - \frac{1}{n}\sum_j^n\left(\alpha\mathbb{E}[\delta_{T,j}]\right) \\
&=& \mathbb{E}[\phi_{T-1,i}] - \frac{1}{n}\sum_j^n\left(\mathbb{E}[\phi_{T-1,j}]\right) \\ 
&&- \mathbb{E}[\phi_{T,i}] + \frac{1}{n}\sum_j^n\left(\mathbb{E}[\phi_{T-1,j}]\right) = 0.
\end{eqnarray}
Thus the property holds for step \(T+1\), and by proof by induction it holds for all \(T\) since the first step holds.

\begin{table}[]
    \centering
        \caption{Summary of final PPL numbers from Reddit data with varying global batch size. All results are from the medium model size, 64 accelerators, four pipeline stages, and data parallel world size of sixteen. }
    \label{tab:ablations}
    \begin{tabular}{l | r  r }
    Method & 1M & 2M \\
    \hline
    FSDP & $19.6$ & $18.0$ \\
    DiLoCo  & $21.0$ & $19.7$ \\
    NoLoCo & $20.9$ & $19.3$ \\
    \end{tabular}
\end{table}

\section{Training losses}
\label{appendix:training_loss}

\begin{figure}[ht]
    \centering
    \begin{subfigure}[t]{0.3\linewidth}
        \centering
        \includegraphics[height=3.8cm]{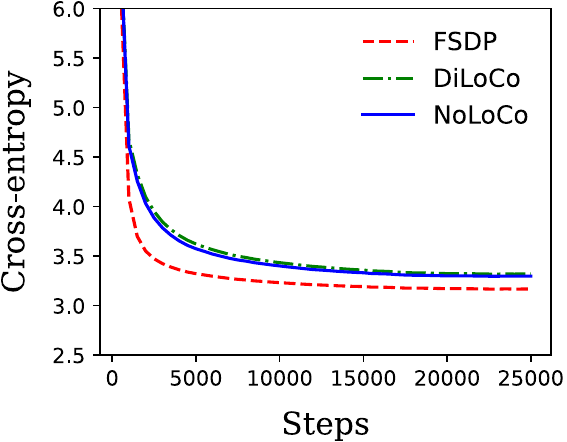}
        \caption{\hspace*{-1.6cm}}
        \label{fig:training_loss_results:A}
    \end{subfigure}
    \hspace{0.3in}
    \begin{subfigure}[t]{0.3\linewidth}
        \centering
        \includegraphics[height=3.8cm]{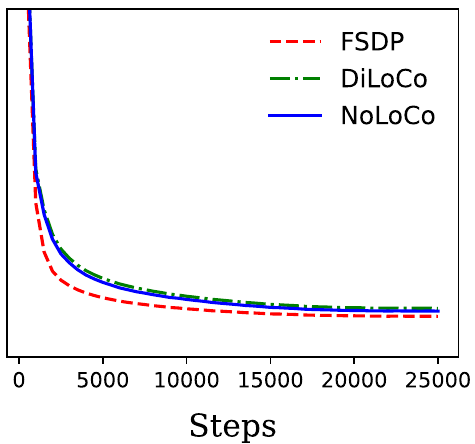}
        \caption{}
        \label{fig:training_loss_results:B}
    \end{subfigure}
    \hspace{-0.02in}
    \begin{subfigure}[t]{0.3\linewidth}
        \centering
        \includegraphics[height=3.8cm]{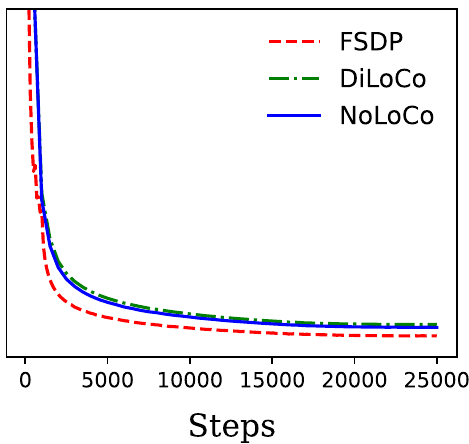}
        \caption{}
        \label{fig:training_loss_results:C}
    \end{subfigure}
    \caption{Reddit training cross-entropy of different training methods at different optimizer step counts. Solid blue curve is NoLoCo, dashed red curve show FSDP, and dashed green curve is DiLoCo. $(a)$: Small model with data parallel world size of $8$ and two pipeline stages; $(b)$: Medium model with data parallel world size of $8$ and two pipeline stages. $(c)$: Large model with data parallel world size of $16$ and four pipeline stages.
    }
    \label{appendix:fig:training_loss}
\end{figure}

Figure \ref{appendix:fig:training_loss} shows training data cross-entropy for all methods and model sizes on Reddit data. Training cross-entropy is averaged over consecutive $200$ (inner) training steps to reduce noise in the data. 

\section{Latency analysis}
\label{appendix:latency}
The approximation $t_{\text{all}} \approx 2 t_c \log_2(n)$ in Section~\ref{sec:th_speedup} assumes uniform communication time. In practice, per-message latency varies, especially on public networks. We model the communication time as $t \sim \mathrm{LogNormal}(\mu, \sigma^2)$. The time for a parent node to receive messages from both children is $t_{\text{local}} = \max(t_1, t_2)$, and for i.i.d.\ log-normal $t_1, t_2$,
\begin{equation}
\mathbb{E}(t_{\text{local}}) = \left(1 + \mathrm{erf}\!\left(\tfrac{\sigma}{2}\right)\right) \exp\!\left(\mu + \tfrac{\sigma^2}{2}\right),
\end{equation}
so the expected local-averaging time is $2 \mathbb{E}(t_{\text{local}})$. This yields the ratio of expected times plotted in Fig.~\ref{fig:latency:A}.

A second source of overhead is that workers do not reach the communication call simultaneously. To isolate this effect, we model each inner step's latency as $\mathrm{LogNormal}(\mu = 1, \sigma^2 = 0.5)$ and measure the time for all processes to finish $250$ outer iterations, ignoring all-reduce and local-averaging time. The resulting ratio is plotted in Fig.~\ref{fig:latency:B}. Reducing the number of inner steps per outer step amplifies this overhead, since the global synchronization point is hit more often.


\end{document}